\documentclass[journal]{IEEEtran}

\ifCLASSINFOpdf
\usepackage[utf8]{inputenc}   
\usepackage[T1]{fontenc}      
\usepackage[pdftex]{graphicx}
\usepackage[table]{xcolor}
\usepackage{array}
\usepackage{listings}
\usepackage{xcolor}
\usepackage{amsmath}
\usepackage{makecell}
\usepackage{orcidlink}
\usepackage{amssymb}
\usepackage{booktabs}
\usepackage{multirow}
\usepackage{ifthen}

\newcommand{\cmark}[1]{\ifthenelse{\equal{#1}{Yes}}{\cellcolor{green!30}Yes}{\cellcolor{red!30}No}}

\lstdefinelanguage{yaml}{
    keywords={configurations, commands, descriptions, Entity, add, type, module_folder, Processor, Initializer, Component},
    keywordstyle=\color{cyan}\bfseries,
    identifierstyle=\color{orange},
    basicstyle=\ttfamily\scriptsize, 
    sensitive=false,
    comment=[l]{\#},
    commentstyle=\color{gray}\itshape,
    stringstyle=\color{red},
    morestring=[b]',
    morestring=[b]",
    literate=
      *{0}{{{\color{green!60!black}0}}}{1}
       {1}{{{\color{green!60!black}1}}}{1}
       {2}{{{\color{green!60!black}2}}}{1}
       {3}{{{\color{green!60!black}3}}}{1}
       {4}{{{\color{green!60!black}4}}}{1}
       {5}{{{\color{green!60!black}5}}}{1}
       {6}{{{\color{green!60!black}6}}}{1}
       {7}{{{\color{green!60!black}7}}}{1}
       {8}{{{\color{green!60!black}8}}}{1}
       {9}{{{\color{green!60!black}9}}}{1}
       {:}{{{\color{orange}:}}}{1}
       {,}{{{\color{orange},}}}{1}
       {-}{{{\color{gray}-}}}{1}
       {true}{{{\color{green!60!black}true}}}{4}
       {false}{{{\color{green!60!black}false}}}{5}
       {no}{{{\color{green!60!black}no}}}{2}
}

\begin{document}

\title{PyGemini: Unified Software Development towards Maritime Autonomy Systems}

\author{Kjetil~Vasstein \orcidlink{0000-0003-4993-3146}, Christian Le \orcidlink{0009-0002-9836-4680}, Simon Lervåg Breivik \orcidlink{0009-0004-5391-4313}, Trygve Maukon Myhr \orcidlink{0009-0005-7066-9644},\\

Annette Stahl \orcidlink{0000-0002-8422-1091}, Edmund Førland Brekke \orcidlink{0000-0001-8735-1687},\\~

Norwegian University of Science and Technology (NTNU), \\
Department of Engineering Cybernetics, \\
O. S. Bragstads plass 2D, 7032 Trondheim, Norway
}

\maketitle

\begin{figure}
    \centering
    \includegraphics[width=\linewidth]{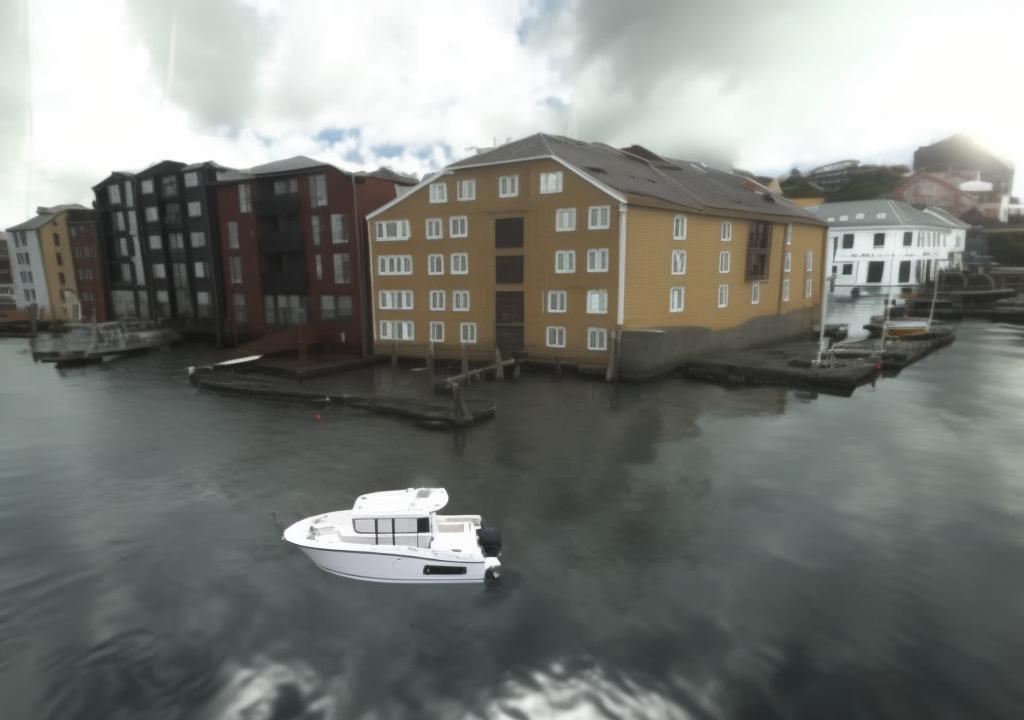}
\end{figure}
\begin{abstract}
Ensuring the safety and certifiability of autonomous surface vessels (ASVs) requires robust decision-making systems, supported by extensive simulation, testing, and validation across a broad range of scenarios. However, the current landscape of maritime autonomy development is fragmented---relying on disparate tools for communication, simulation, monitoring, and system integration---which hampers interdisciplinary collaboration and inhibits the creation of compelling assurance cases, demanded by insurers and regulatory bodies. Furthermore, these disjointed tools often suffer from performance bottlenecks, vendor lock-in, and limited support for continuous integration workflows. To address these challenges, we introduce PyGemini, a permissively licensed, Python-native framework that builds on the legacy of Autoferry Gemini to unify maritime autonomy development. PyGemini introduces a novel Configuration-Driven Development (CDD) process that fuses Behavior-Driven Development (BDD), data-oriented design, and containerization to support modular, maintainable, and scalable software architectures. The framework functions as a stand-alone application, cloud-based service, or embedded library---ensuring flexibility across research and operational contexts. We demonstrate its versatility through a suite of maritime tools---including 3D content generation for simulation and monitoring, scenario generation for autonomy validation and training, and generative artificial intelligence pipelines for augmenting imagery---thereby offering a scalable, maintainable, and performance-oriented foundation for future maritime robotics and autonomy research.
\end{abstract}


\section{Introduction}
\IEEEPARstart{E}{nsuring} the safety of autonomous surface vessels (ASVs) is paramount to their regulatory acceptance and commercial deployment---particularly as insurers and classification societies demand robust assurance cases before approving such systems for operation. Achieving this assurance requires extensive testing across a wide range of scenarios, failure modes, and environmental conditions, many of which are impractical, time-consuming or unsafe to evaluate at sea. Simulation has thus become an indispensable tool in the maritime domain for demonstrating functional safety, verifying system behavior, and supporting certification efforts \cite{Glomsrud:Modular_assurance, Torben:contract_based_verification}. In parallel, maritime autonomy is becoming increasingly data-driven, as AI-enabled systems for perception, situational awareness (SA), navigation, and decision-making rely heavily on large-scale, high-quality training datasets. This shift amplifies the demand not only for safety assurance but also for scalable data generation pipelines. However, there has been a lack of specialized simulation platforms capable of addressing the unique challenges of ASVs---such as high-fidelity sensor modeling, complex environmental interactions, and integration with real autonomy software stacks.

As a response to this gap, Autoferry Gemini was developed as one of the first maritime extensions of a game engine, enabling GPU-based sensor models tailored for maritime applications \cite{Vasstein_autoferry_gemini}. Since its inception, this toolset has supported a wide range of technical \cite{Vagale:navigation_evaluation, Vasstein:Hellinger, Torben:contract_based_verification}, systemic \cite{Cheng:Human_errors, Hoem:criop} and design \cite{Veitch:human_centered_explainable_AI, Veitch:study_of_asv_operator_attention} oriented research efforts toward ASVs, in addition to motivating creation of similar extensions for underwater robotics \cite{Loncar:Marus}. Collectively, these efforts have played a key role in the development and operation of the ASVs milliAmpere 1 and 2 \cite{Brekke:milliAmpere, Eide:milliampere2}, as well as their accompanying shore control center \cite{Alsos:Shore_control_labl}.

Given the breadth of these applications and emerging demands, such as handling licensing issues for 3D assets and cybersecurity requirements for new ships \cite{Zafar:cybersecurtiy_demands_requirements}, there is a clear need for a more advanced and cohesive tool that can support both current and future research in maritime autonomy. Moreover, since the majority of R\&D activities surrounding the autonomy systems are conducted in Python \cite{Brekke:VIMMJIPDA, Nygård:fast_sam}, a next-generation solution should tightly integrate Python to effectively bridge the technical, systemic, and design-oriented research mentioned above.

Building on these insights, and continuing the work initiated by Autoferry Gemini \cite{Vasstein_autoferry_gemini}, we present \textit{PyGemini}---a permissively licensed, Python-native framework designed to unify maritime autonomy development.
To better maintain the broader use cases than existing solutions, we develop a new novel Configuration-Driven Development (CDD) workflow that combines containerization \cite{Merkel:Docker}, Entity-Component-System (ECS) architecture \cite{capdevila:ECS_in_serious_games}, and Behaviour-Driven Development (BDD) \cite{Solis:BDD_charachteristics} to promote modular, reprocessable, and collaborative research.
With emphasis on either complementing or replacing existing solutions, PyGemini offers the flexibility to run as either a stand-alone application, service in the cloud, or library in external contexts. Key contributions of this paper include:
\begin{itemize}
\item A unified, open-source framework for maritime application development using Python.
\item A novel development process (CDD) combining ECS and BDD principles for maintainable development.
\item Several relevant maritime autonomy applications and a growing repository of open licenced maritime assets \cite{Vasstein:PyGemini_Data}.
\end{itemize}

We begin by introducing related work and giving a comparison between current solutions and PyGemini. We continue outlining PyGemini’s architecture and design principles, followed by an in-depth look at how these inform the CDD process. We then demonstrate several technical subsystems in use cases, before discussing PyGemini in context of existing solutions. Finally, we conclude the paper with a summary and propose a way forward.

\section{Related Work}
In software development, libraries, frameworks, and platforms form a hierarchical ecosystem of software architectures, each serving distinct roles at increasing abstraction levels. Libraries provide reusable code modules for specific tasks, frameworks offer structured workflows for building applications, and platforms serve as comprehensive ecosystems for deploying and running applications.

\begin{figure}[htbp]
    \centering
    \includegraphics[width=1.0\linewidth]{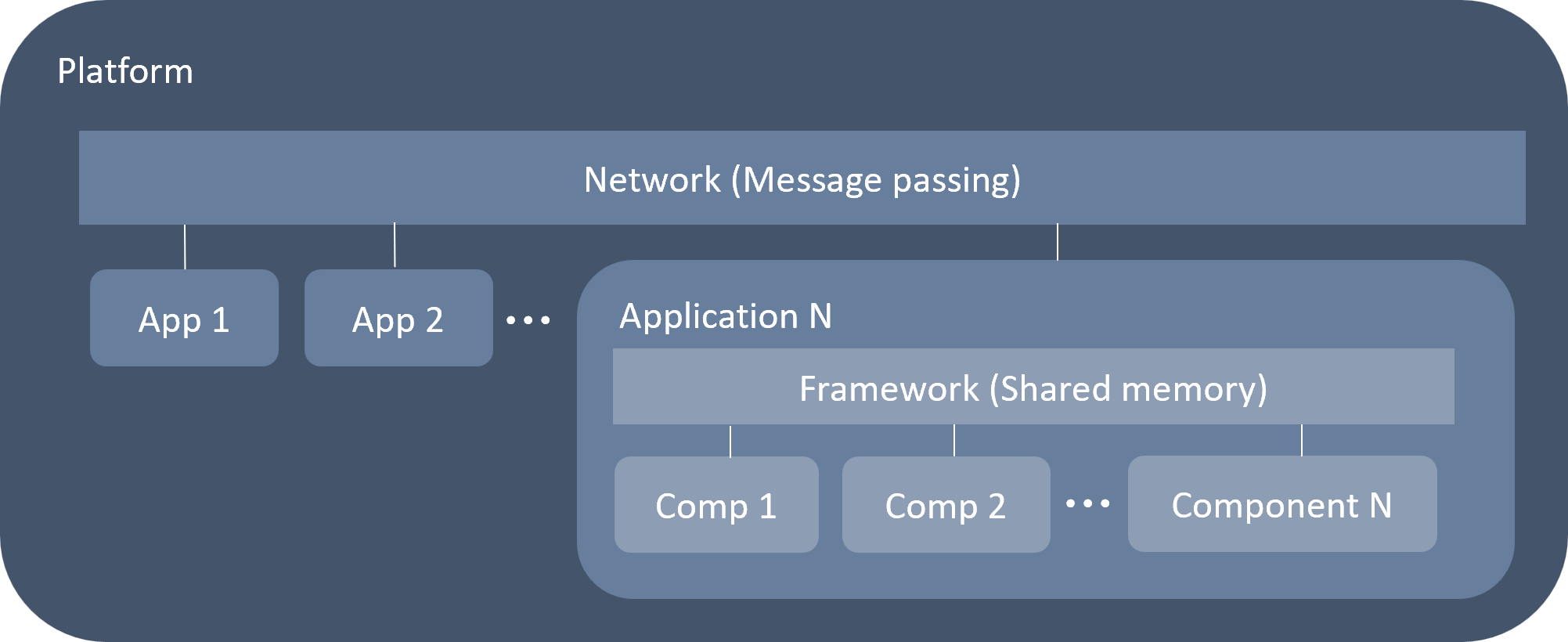}
    \caption{Typical layer setup in platform-based systems. Platform and application layers may use similar architectural patterns but typically differ in naming conventions and communication protocols.}
    \label{fig:software_layers}
\end{figure}

PyGemini focuses on addressing the application layer (Figure \ref{fig:software_layers}), which serves as the cornerstone in any development as it involves both libraries and frameworks. We therefore give a broad summary of related work found in existing platforms, frameworks, and libraries of relevance to maritime autonomy. We highlight how each addresses, or fails to address---critical requirements for unified maritime autonomy development, providing a qualitative comparison in Table \ref{tab:feature_comparison} as a summary of how existing solutions fair against PyGemini.

\begin{table*}[htbp]
\centering
\renewcommand{\arraystretch}{1.2}
\setlength{\tabcolsep}{8pt}
\begin{tabular}{llcccccccc}
\toprule
\textbf{} & \textbf{} & \multicolumn{2}{c}{\textbf{Platforms}} & \multicolumn{2}{c}{\textbf{Engines}} & \multicolumn{3}{c}{\textbf{Engine extensions}} & \textbf{PyGemini} \\
\cmidrule(lr){3-4} \cmidrule(lr){5-6} \cmidrule(lr){7-9}
\textbf{Category} & \textbf{Characteristic} & \textbf{ROS} & \textbf{OSP} & \textbf{Vico} & \textbf{Stonefish} & \makecell{\textbf{HoloOcean} \\ \textbf{(Unreal)}} & 
\makecell{\textbf{MARUS} \\ \textbf{(Unity)}} & 
\makecell{\textbf{Autoferry Gemini} \\ \textbf{(Unity)}} \\
\midrule
\multirow{3}{*}{Communication} 
& Message-passing             & \cmark{Yes} & \cmark{Yes} & \cmark{Yes} & \cmark{Yes} & \cmark{Yes}  & \cmark{Yes}  & \cmark{Yes}  & \cmark{Yes}\\
& Shared-memory               & \cmark{No}  & \cmark{No}  & \cmark{Yes} & \cmark{Yes} & \cmark{Yes}  & \cmark{Yes}  & \cmark{Yes}  & \cmark{Yes}\\
& GPU-GPU                     & \cmark{No}  & \cmark{No}  & \cmark{Yes} & \cmark{Yes} & \cmark{Yes}  & \cmark{Yes}  & \cmark{Yes}  & \cmark{Yes}\\
\midrule
\multirow{7}{*}{Simulation}
& Adjustable Fidelity         & \cmark{No}  & \cmark{No}  & \cmark{No}  & \cmark{No}  & \cmark{Yes} & \cmark{Yes}  & \cmark{Yes}  & \cmark{Yes}\\
& Camera                      & \cmark{No}  & \cmark{No}  & \cmark{Yes} & \cmark{Yes} & \cmark{Yes} & \cmark{Yes}  & \cmark{Yes}  & \cmark{Yes}\\
& Lidar                       & \cmark{No}  & \cmark{No}  & \cmark{No}  & \cmark{No}  & \cmark{No}  & \cmark{No}   & \cmark{Yes}  & \cmark{Yes}\\
& Sonar                       & \cmark{No}  & \cmark{No}  & \cmark{No}  & \cmark{Yes} & \cmark{Yes} & \cmark{Yes}  & \cmark{No}   & \cmark{No}\\
& Radar                       & \cmark{No}  & \cmark{No}  & \cmark{No}  & \cmark{No}  & \cmark{No}  & \cmark{No}   & \cmark{Yes}   & \cmark{No}*\\
& Motion Physics              & \cmark{No}  & \cmark{No}  & \cmark{Yes} & \cmark{Yes} & \cmark{Yes} & \cmark{Yes}  & \cmark{Yes}   & \cmark{No}*\\
& IP-protection               & \cmark{No}  & \cmark{Yes} & \cmark{No}  & \cmark{No}  & \cmark{No}  & \cmark{No}   & \cmark{No}   & \cmark{No}\\
\midrule
\multirow{4}{*}{Development}
& Python support              & \cmark{Yes} & \cmark{Yes} & \cmark{No}  & \cmark{No}  & \cmark{Yes} & \cmark{No}  & \cmark{No}  & \cmark{Yes}\\
& Portable libraries          & \cmark{Yes} & \cmark{Yes} & \cmark{Yes} & \cmark{Yes} & \cmark{No}  & \cmark{No}  & \cmark{No}  & \cmark{Yes}\\
& Permissive license          & \cmark{Yes} & \cmark{No}  & \cmark{No}  & \cmark{No}  & \cmark{No}  & \cmark{No}  & \cmark{No}  & \cmark{Yes}\\
& BDD/CDD                     & \cmark{No}  & \cmark{No}  & \cmark{No}  & \cmark{No}  & \cmark{No}  & \cmark{No}  & \cmark{No}  & \cmark{Yes}\\
\bottomrule
\end{tabular}
\newline
\caption{Comparison of characteristics across platforms and frameworks. Green and red indicate presence and absence, respectively.\\
* Currently in development or planned for a future release}
\label{tab:feature_comparison}
\end{table*}

\subsection{Platforms}
Software platforms have long served as foundational tools for collaboration across robotics and autonomy. Among the most widely adopted in the maritime domain have been the Robot Operating System (ROS) \cite{Quigley:ROS}, MOOS-IvP \cite{Benjamin:MOOS_ivp} and LSTS \cite{Pinto:LSTS}, which provide standardized middleware for communication between heterogeneous software modules and hardware systems. Research ASVs such as the milliAmpere ferries use ROS to build modular navigation systems for SA and decision-making \cite{Brekke:milliAmpere, Eide:milliampere2}. These systems include sensor fusion, perception, and tracking modules for interpreting dynamic elements in the environment \cite{Brekke:VIMMJIPDA}. Outputs from these SA modules feed into motion planning and collision avoidance systems, which interface with motion control to guide vessel actuators.

The systems described above typically integrate proprietary hardware and software from multiple vendors in custom configurations. Given the safety-critical nature of maritime operations, rigorous testing of component interoperability is vital. However, real-world testing is often constrained by cost, safety, and environmental unpredictability. As a result, Hardware-in-the-Loop (HIL) testing has become a standard approach to validate functionality, performance, and fault tolerance using virtual test beds in controlled environments \cite{Pedersen:HIL, Smogeli:HIL}. The increasing complexity of autonomous systems, however, demands broader test scopes that exceed what HIL alone can offer \cite{Glomsrud:Modular_assurance}. This is especially important to prevent or spot adversarial cyberattacks on machine learning supply chains, datasets or similar that leverages the weaknesses of complex systems \cite{Walter:AAI_maritime_testcases}. HIL setups are also limited by the real-time execution requirements of physical hardware. While ROS, MOOS-IvP, and LSTS support HIL configurations by interfacing with external simulators, they lack integrated simulation environments and provide limited Intellectual Property (IP) protection for vendor models during testing.

To address these limitations, the maritime sector---led in joint effort between DNV, Kongsberg, Sintef and NTNU---developed the Open Simulation Platform (OSP) \cite{Smogeli:OSP}, which leverages the Functional Mock-up Interface (FMI) standard to enable IP protected co-simulations across vendor-specific modules \cite{Blochwitz:FMI}. Within OSP, Software-in-the-Loop (SIL) testing expands the test scope by removing physical hardware constraints, enabling scalable validation in cloud environments. SIL configurations are also adaptable for integration into HIL setups, facilitating seamless transitions between testing stages. In such heterogeneous environments, containerization platforms such as Docker play a key role by encapsulating software dependencies, standardizing deployment, and improving reproducibility across diverse systems \cite{Merkel:Docker}. This consistency is especially valuable for platforms such as OSP, where interoperability and data integrity must be maintained across modular components.

While SIL and containerization enhance reproducibility and scalability, the message-passing architectures underlying platforms such as ROS, MOOS-IvP, LSTS, Docker, and OSP introduce significant throughput bottlenecks---particularly in GPU-intensive workflows. These bottlenecks stem from frequent CPU-GPU memory transfers, which degrade real-time performance in both simulations and onboard autonomy. Situational awareness (SA) modules are especially impacted, as high-throughput sensor streams from either onboard sensors or simulation models (e.g., lidar, radar, and cameras) \cite{Wagner:static_radar_equation, Vasstein_autoferry_gemini, Kerbl:3Dgaussians} must be transferred across operating system (OS) processes on either CPU or GPU instances. While storing such data in formats like ROS bags enables offline or open-loop applications such as AI training, this approach incurs high storage costs and is unsuitable for closed-loop autonomy testing and operation due to the lack of dynamic feedback. Hardware-level optimizations such as NVIDIA’s GPUDirect RDMA offer more direct data pathways, but these solutions are hardware-specific and not broadly supported \cite{Hamidouche:GPU_direct_RDMA}. As more simulation and autonomy models migrate to GPU execution, these platform-level bottlenecks are likely to intensify---further limiting real-time capabilities of autonomy systems and achievable test coverage that could validate them.

These limitations largely stem from OS-level constraints designed to ensure process isolation and program safety. Additionally, the decentralized structure of platforms often leads to duplicated codebases and increased maintenance complexity. Nonetheless, platforms offer a robust foundation for interdisciplinary development by adopting principles such as inversion of control, allowing individual contributors to build within a shared software architecture without centralized coordination.

\subsection{Frameworks}
Similar architectural patterns used in the platform layer, also exist in the application layer (Figure~\ref{fig:software_layers}). Here, shared-memory communication often replaces network-based message passing, enabling higher throughput via framework systems. Traditionally, many engines---such as those used in creating video games---employed object‑oriented frameworks, particularly when their scope was limited to tasks like graphics rendering \cite{Tancik:nerfstudio, Zhou:open_3d}. However, modern engines with broader functionality increasingly adopt data‑oriented designs such as the ECS. ECS has proven effective in maritime co‑simulations \cite{HATLEDAL:Vico} and advanced simulators \cite{Zhou:Isaac_sim}, but have been most widely adapted in proprietary mainstream game engines such as Unity and Unreal. By organizing data for cache‑friendly access and parallel processing, these frameworks alleviate GPU and CPU memory bottlenecks experienced in platforms, enabling high‑fidelity, real‑time simulations without sacrificing modularity.

However, proprietary engines pose risks such as vendor lock-in and limited long-term support, often due to restrictive licensing. Unity’s discontinuation of its cloud-based simulation platform in December 2023 exemplifies these vulnerabilities, leaving adopters without access to previously hosted systems and source code \cite{unitySimulationDeprecation2023, bloombergUnityRefocus2024, reutersUnityLayoffs2024}.

Open-source alternatives such as Vico \cite{HATLEDAL:Vico}, UUVsim \cite{Zhang:UUVSim}, UWsim\cite{Prats:UWsim} and Stonefish \cite{Cieslak:Stonefish, grimaldi:stonefish} mitigate some of these concerns but remain constrained by licensing terms. For instance, UWsim and Stonefish have GNU General Public Licenses (GPL), which requires that derivative works disclose their source code, discouraging proprietary integration \cite{Valimaki_licencing_evaluation}. Vico, while publicly available, does not specify an explicit license at the time of writing, which restricts its legal reuse or integration under standard open-source practices. In contrast, permissive licenses such as MIT and Apache 2.0 used by UUVsim are more commercially attractive, imposing minimal obligations on redistribution \cite{Alspaugh_licencing_roles, Santos:licencing_strategies}. However, UUVsim focuses on underwater simulations and does not seem to be under active development. Few relevant open-source engines are therefore available under permissive licensing terms. 

Similar challenges arise with 3D content, where high-quality assets are typically purchased and licensed, restricting sharing and reuse among researchers and the open-source community. This is particularly problematic given that high-fidelity simulation depends not only on accurate models but also on richly detailed content. Moreover, even high-quality assets are often handcrafted, lacking the diversity and variability present in real-world environments. With lack of standards for validating data from simulators \cite{Vasstein:Hellinger}, the influence of 3D content on simulations---and on the research that depends on these simulations---remains unclear.

Despite their use in maritime simulators \cite{Veitch:study_of_asv_operator_attention, Hoem:criop, Vagale:navigation_evaluation, Cheng:Human_errors}, game engines often lack autonomy-specific tools necessary for creating digital twins. Third-party extensions, such as Autoferry Gemini \cite{Vasstein_autoferry_gemini}, MARUS \cite{Loncar:Marus}, UNav-Sim \cite{Amer:unav_sim}, and HoloOcean \cite{Potokar:holoocean}, do exist. However, the underlying engines---originally developed for video games---often impose source code restrictions and architectural limitations that hinder development for use cases beyond high-fidelity simulations. These include low-fidelity simulations, integration with autonomy systems, remote monitoring, machine learning workflows, and cybersecurity test beds. Middleware solutions such as ROS can bridge some of these gaps, but often at the expense of performance and portability. Additionally, engines typically use performance-oriented languages such as C\# or C++, often with limited access to external libraries, which confines development within the engine environment.

This poses a notable constraint in research and development, where recent breakthroughs in autonomy and simulation---such as Segment Anything Models (SAM) for docking \cite{Ravi:sam2, Nygård:fast_sam}---scenario generation from smart agents \cite{Fjellheim:Agents} or probabilistic programming \cite{Fremont:scenic}---and novel view synthesis (NVS) techniques---such as 3D Gaussian Splatting (3DGS) \cite{Kerbl:3Dgaussians}, Neural Radiance Fields (NeRF) \cite{Mildenhall:NeRF}, and Stable Diffusion (SD) \cite{Rombach:Latent_diffusion} with ControlNets \cite{zhang:ControlNets}---are predominantly built using open-source, Python-native libraries for CPU and GPU computation, including NumPy \cite{harris:NumPy}, PyTorch \cite{Ansel:Pytorch2}, and Numba \cite{Lam:Numba}. These libraries have also narrowed the performance gap with other languages, making Python increasingly viable for high-performance tasks. Furthermore, the availability of packages from Nerfstudio \cite{Tancik:nerfstudio, Ye:gsplat} for 3DGS and NeRF generation and visualization, and Diffusers \cite{Von_platen:diffusers} for image augmentation, opens up novel approaches to address licensing and accessibility issues related to 3D content. In Section \ref{sec:use_cases} we will demonstrate this and other use cases dependent on the aforementioned libraries.

\subsection{Software Development}
We have so far seen that current approaches to developing autonomy and supporting tools are often fragmented---confined within isolated libraries, frameworks, and platforms. This siloed structure hinders interdisciplinary research and integration, leading to inefficiencies in scalability, data management, and overall system design. Additionally, a heavy reliance on proprietary tools and licensed content---paired with weak or restrictive open-source alternatives---complicates long-term maintenance, reproducibility, and collaborative development. The result is redundant effort, inflated complexity, and slow progress toward truly integrated systems.

To address these systemic issues, it is essential to consider software architectures and development methodologies that inherently promote modularity, interoperability, and collaboration. In addtion of providing reprodusability across divserse system, containerisation with docker allows creating unified work environments that reduces configuration drift and facilitates collaboration among developers, testers, and researchers \cite{Merkel:Docker}. Beyond technical efficiency, the ECS paradigm has demonstrated its value as a scalable framework for large-scale software development both in the video game industry and for serious games \cite{capdevila:ECS_in_serious_games}. By fostering modular design and a clear separation of concerns, ECS supports coordination across interdisciplinary teams. In parallel, Behavior-Driven Development (BDD) has been employed as a process to further align collaboration in research frameworks \cite{Binamungu:BDD_Map}. Here a \textit{ubiquitous language} is established to specify and translate user requirements into acceptance tests to better align user expectations with developer goals \cite{Solis:BDD_charachteristics}. However, effectively applying BDD can be challenging in large-scale projects, particularly when evolving requirements complicate behaviour specification and increase maintenance costs, while distributed teams face difficulties in managing version control and system-level test cases \cite{Irshad:BDD_challenges}. To address these issues, \cite{Irshad:BDD_challenges} recommends defining a system-level feature file early in development to unify terminology and serve both as a requirement specification and an automated system test. We explore PyGemini's approach to this in Section \ref{sec:architecture}

\section{PyGemini} \label{sec:architecture}
To enable scalable and collaborative development across disciplines, PyGemini adopts a CDD approach that integrates BDD principles with an ECS architecture. In this approach, configuration files act as the system-level feature file---aligned with the concept proposed by \cite{Irshad:BDD_challenges}---becoming a central artifact guiding the entire development process. These files define the application behaviour declaratively using ECS terminology---becoming the \textit{ubiquitous language} in BDD. In our approach, this allows the configuration files to serve multiple purposes:
\begin{itemize}
    \item They are entry points and examples for new users to base their application from.
    \item They serve as a specification for the application build, determining the application’s behaviour.
    \item They serve as user requirements to PyGemini, manifested as acceptance tests.
\end{itemize}
This ensures that user intent is consistently embedded into the software lifecycle, while keeping maintenance low.

The files also separates concerns through a front-end/back-end structure: the back-end contains the ECS framework implementation and core libraries, while the front-end comprises configuration files that specify which components, processors, and entities to use for building a given application. These configuration files provide a stable, higher-level language that abstracts underlying implementation details and enables users to construct applications naturally using the ECS vocabulary.

In the following subsections, we first introduce the back-end system consisting of PyGemini's ECS framework and library. We then show how configuration files are authored and used to build applications, then explain how they integrate with the testing system. Finally we illustrate how all this plays an integral role in CDD, before showing some standard components used by the applications in the next section.

\subsection{ECS Framework} \label{sec:ECS}
ECS promotes a modular and declarative design philosophy, where behaviour emerges from the interaction of independent data and logic units, rather than being encoded in rigid class hierarchies. This approach encourages separation of concerns, improves reusability, and aligns naturally with data-driven development--particularly advantageous in simulations, game development, and other dynamic modelling environments. To avoid confusion with terminologies, we break down the ECS paradigm into its three foundational elements: Entities, Components, and Processors:

\textbf{Entities} are unique identifiers that act as pointers to data stored in components. They do not hold data themselves but represent objects through their assigned components. This design follows the principle of composition, famously encapsulated by the \textit{Duck Test} proverb often associated with abductive reasoning: “If it looks like a duck, swims like a duck, and quacks like a duck, it probably is a duck.” Thus, an entity’s functionality is determined by the components it holds, ranging from tangible objects to abstract concepts like time.

\textbf{Components} define the data structure of the application, serving as containers for specific attributes. In PyGemini, components are implemented as data classes, adhering to Python’s native types and standard libraries such as NumPy. This standardization gives components the second role to work as interfaces, facilitating integration between libraries and improving code scalability.

\textbf{Processors} handle the logic and interactions between components. Each processor performs a specific function, and the combined behavior of multiple processors creates the application’s emergent behavior. Ideally, processors only handle data flow and interaction, delegating functional logic to the library’s core functions.

\subsection{Library} \label{sec:library}
The library in PyGemini encapsulates core functionalities that define what actions occur within the ECS framework. This includes logic for AI models, sensor simulations, and data processing routines. When integrated with the ECS framework, the library’s core functions determine how component data is updated based on interactions between entities.

A distinct type of functions, \textbf{Initializers}, handles the setup of large components, such as mesh data from 3D files. These functions operate primarily at the start of the application, allowing for flexible attribute initialization -- such as downloading a mesh file from a server, generating procedural geometry, or customising sensor characteristics such as a lidar's beam pattern. This modular approach to attribute assignment enhances composability and adaptability within PyGemini’s architecture.

\subsection{Configuration file} \label{sec:configuration_file}
\begin{figure*}[!ht]
    \centering
    \begin{minipage}{0.4\textwidth}
\begin{lstlisting}
configurations:
  - add: content/marina_model.yaml
commands:
  Processor:
    type: processor
    module_folder: pygemini.processors
  Component:
    type: component
    module_folder: pygemini.components
  Initializer:
    type: function
    module_folder: pygemini.initializers.mesh
  Entity:
    type: entity
descriptions:
  - Processor: UpdateRenderingScene
    priority: 100000
  - Processor: RotateTransforms
    priority: 10004
  - Processor: UpdateMesh
    priority: 10003
  - Processor: UpdateTime
  - Processor: UpdateCameras
  - Entity:
    - Component: RenderingScene
  - Entity:
    - Component: Time
      timestep: 0.5
  - Entity:
    - Component: Camera
      width: 512
      height: 512
      focal_length: 0.8
  - Entity:
    - Component: BaseMesh
      mesh:
        Initializer: create_coordinate_frame
        size: 100
      is_dynamic: no
    - Component: Transform
\end{lstlisting}
    \end{minipage}
    \hspace{1cm}
    \begin{minipage}{0.38\textwidth}
        \centering
        \includegraphics[width=\linewidth]{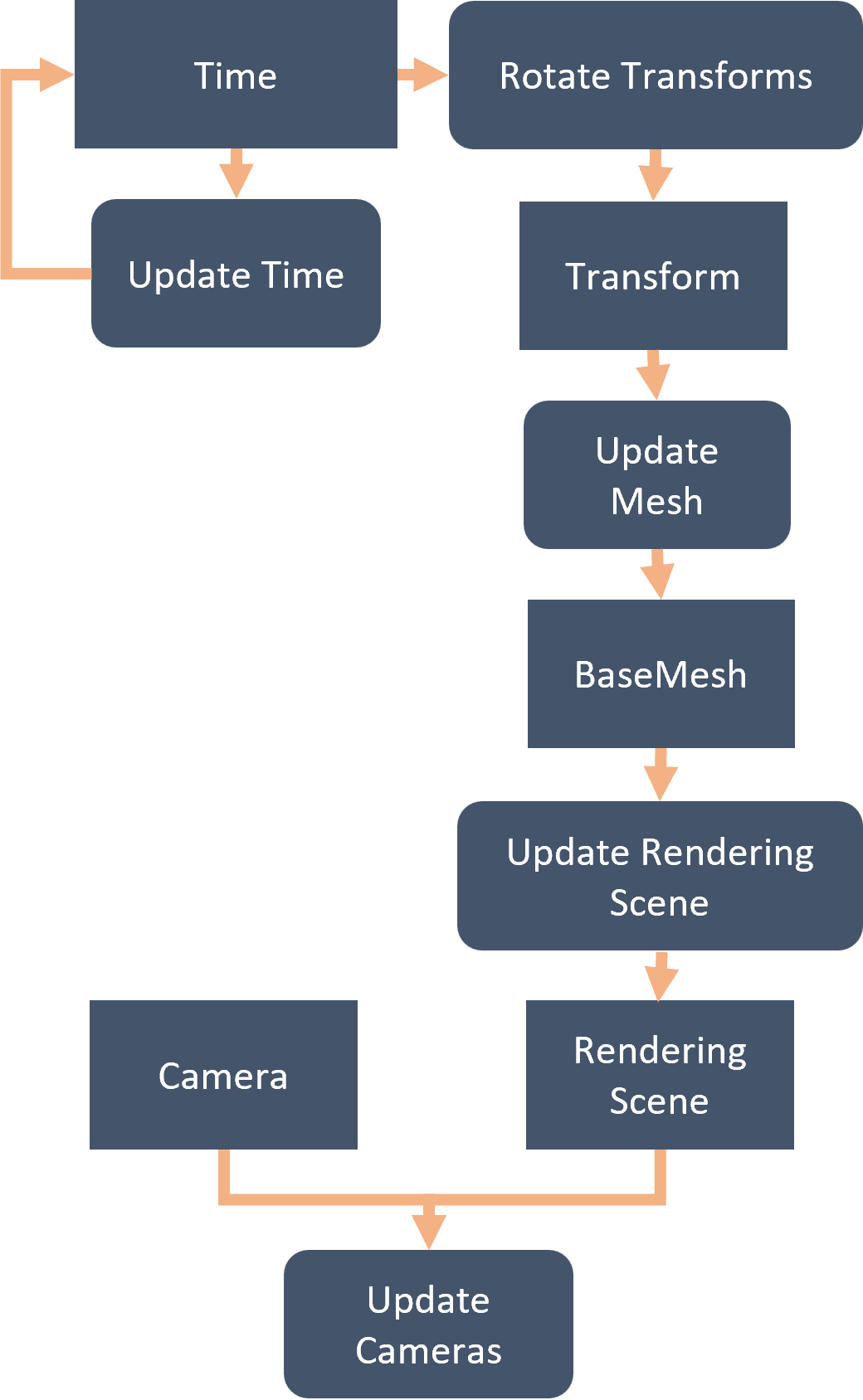}
    \end{minipage}
    \caption{Example of a PyGemini configuration file to the left, and the resulting data pipeline to the right. Regular rectangles are components, while rounded are processors.}
    \label{fig:configuration_and_datapipeline}
\end{figure*}

PyGemini’s framework allows several applications to be built according to user specifications. Some of the more advanced applications will be shown in the results section, while here we focus on the basics of how these are made from configuration files.

In the left portion of Figure \ref{fig:configuration_and_datapipeline} we have made a short example of a configuration file that renders a coordinate frame rotating around its axis. A more visual representation can be seen in the figure's right portion, displaying a data pipeline. This shows how data will flow between components as a result of processors.

The pipelines often form tree structures but also loops, such as in the case with the "Update Time" processor in Figure \ref{fig:configuration_and_datapipeline}. These are built by commands defined in the configuration file's "commands" section (Figure \ref{fig:configuration_and_datapipeline}). There are several types of commands, where some require reference to modules in the back-end system to work:
\begin{itemize}
    \item Entity, serves as a placeholder to which components can be attached.
    \item Component, creates a component inside an entity requiring attributes to be set. It needs a path to the module where the component is implemented. The command ``Component'' is an example of a command that uses PyGemini's built-in component module.
    \item Processor, creates a processor that determines how data flows in the application. It needs a path to the module where the processor is implemented. The command ``Processor'' is an example of a command that uses PyGemini’s built-in processor module.
    \item Function, serves as a helper to initialize components attribute values with module functions. It needs a path to the module where the function is implemented. The command ``Initializer\_mesh'' is an example of a command that uses one of the many initializer modules in PyGemini’s library.
\end{itemize}
The first three commands effectively complete the ECS framework. Commands are called in the configuration’s description section, defining every entity, component and processor for an application. Configurations can also consist of other configuration files by referencing them in the configuration section. This allows configuration code to be re-used to create new data pipelines or to form instances of identical entities, such as vessels and terrain.

Processors can easily be added and removed, which is useful in several circumstances. A camera could for instance be used for debugging purposes before final cloud deployment where it is not needed. The same configuration file could be used by just commenting out the processor. Interchanging processors near leaf nodes in the tree structure makes it possible to run from the same application state. Sensor models such as \textit{Update Cameras} in Figure \ref{fig:configuration_and_datapipeline} that renders to screen could from this be interchanged to vary sensor fidelity levels. As an example, Figure \ref{fig:scenario_visualization} demonstrates how various rendering models, based on the same component data, could provide different scene appearances.

To build and run the configuration, a simple Python script can be written, referencing the path to the configuration file. This builds the application as a "world" which could either be run directly or be further edited in Python if needed (e.g. to automate generation of similar applications for cloud services).

\subsection{Test system}
After a user is satisfied with their application, the configuration file can optionally be submitted as an acceptance test to ensure compatibility with future versions of PyGemini. From PyGeminis standpoint, the configuration now becomes user requirements that must be met for any future update, i.e. an acceptance test. The test is generated by logging binary data from every component attribute on every entity for every iteration in the application. We will refer to this as the \textit{Application’s state}, where an example can be seen in Table \ref{tab:application_state} .

To save storage we choose to only log states if they have changed since the last iteration. Some of the attribute’s data can also be very large, such as with image data, point clouds or mesh. To further save space, the binary data goes through a SHA-256 hash function giving it a unique and reproducible hash id with a small and fixed storage size. This gives the acceptance test a small storage footprint, making them suitable in version control systems alongside code. 

\begin{table}
    \centering
    \small
        \begin{tabular}{|l|l|c|c|p{1.0cm}|}
        \hline
        \textbf{Entity} & \textbf{Iter.} & \textbf{Component} & \textbf{Attribute} & \textbf{Data} \\
        \hline
        1 & 1 & Transform & world           & 613e… \\
        1 & 1 & Transform & local           & 1c9a… \\
        1 & 1 & Mesh      & tensor          & 187c… \\
        2 & 1 & Time      & currentTime     & 0a02… \\
        2 & 1 & Time      & increment\_step & 3d6c… \\
        1 & 2 & Transform & world           & 24a5… \\
        2 & 2 & Time      & currentTime     & e303… \\
        1 & 3 & Transform & world           & eaeb… \\
        2 & 3 & Time      & currentTime     & 14ee… \\
        \hline
        \end{tabular}
        \newline
    \caption{A section from an application state, showing entity id's, iteration step, components, attributes and hashed attribute data.}
    \label{tab:application_state}
\end{table}

In version control systems, when changes are made to the code base, tests are run by reproducing the application states. If all acceptance tests succeed, there have been no changes in the code that would affect user requirements specified by the configuration files. If the application state has changed, the acceptance test fails. This makes developers aware that they have introduced a potential breaking change for users. Failing the acceptance test could be due to several reasons:
\begin{itemize}
    \item Editing configuration files could have introduced new entities, attached new components or changed their attribute values. If that is the case, the user requirements of the original application have changed, and this should be reflected in updating the acceptance test in coordination with the user.
    \item Back-end code has changed, affecting data stored in component attributes. This could either be intentional or a bug.
    \begin{itemize}
        \item If it is intentional, a code update in collaboration with the user behind the acceptance test should be done. Fails could be due to solving another bug, which means the user’s requirement was based on false premises, or code refactoring such as changing names on attributes. Either way, users must evaluate if their requirements can still be met in the developers update.
        \item If it is a bug, then the developer must solve it before the changes to the code is accepted.
    \end{itemize}
\end{itemize}

The philosophy is that code might change, but the user’s requirement should always be in the loop while doing so. This keeps developers informed on how their code impacts users as the acceptance tests effectively test the integration of components, processors and core functions. Developers can then make measures on how they can prevent integration conflicts when these tests fail. If not, the users should be made aware of how and what must be changed in their requirements.

\subsection{Configuration Driven Development}

\begin{figure}
\centering
\includegraphics[width=1.0\linewidth]{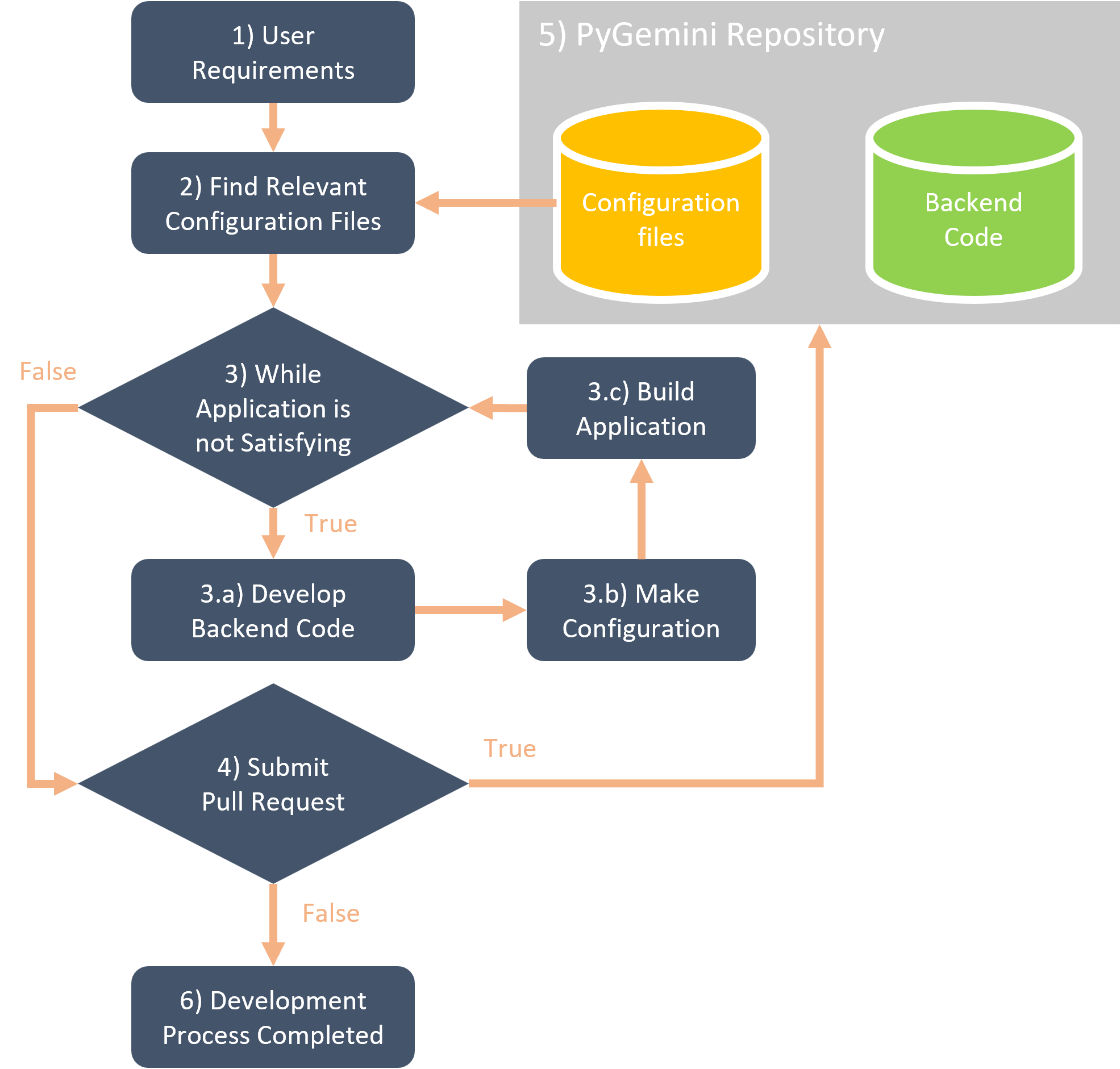}
\caption{Flowchart illustrating the Configuration-Driven Development process in PyGemini.}
\label{fig:cdd_process}
\end{figure}

PyGemini leverages a configuration-driven development (CDD) approach to streamline application development and testing. This approach integrates configuration files as both entry points and test cases, fostering a maintainable and iterative workflow that encourages user contributions. Figure \ref{fig:cdd_process} outlines the CDD workflow, comprising the following steps:
\begin{enumerate}
    \item The process begins with users defining the requirements for their desired application.
    \item Users search the open-source repository for existing configuration files. These configurations serve as entry points, allowing users to build and run pre-existing applications to assess their suitability.
    \item If the existing applications do not fully meet user requirements, users may transition to a developer role, initiating an iterative development cycle:
    \begin{enumerate}
        \item New processors, components, or core library functions may be implemented to address missing features. Existing components provide structural guidelines, while core library functions streamline processor development.
        \item A new configuration file is created or modified based on the user requirements. If new back-end components are introduced, the configuration must integrate these changes to fulfill user requirements.
        \item Once the configuration is complete, the application is built and tested against the user requirements.
    \end{enumerate}
    \item If the user wishes to contribute the developed feature, they may submit a pull request to the PyGemini repository, including both the new configuration file and the corresponding back-end code.
    \item Upon approval, the configuration file is added to PyGemini’s repository, serving as both an entry point for new users and an acceptance test for PyGemini's test system. This ensures that the newly developed feature is maintained in future updates.
    \item Whether or not a pull request is submitted, the development cycle concludes once the application successfully meets the user requirements.
\end{enumerate}

This workflow not only fosters modular and reusable code but also establishes a self-reinforcing development model, where each contribution increases the repository’s utility for future users and developers.

\subsection{Components} \label{sec:components}
In this section, we provide a brief overview of some of PyGemini’s main component systems as they currently stand. While new components can be added, it is generally advisable to reuse existing ones to promote consistent interfaces between processors. Moreover, composing multiple components on an entity---as illustrated in the configuration file in Figure~\ref{fig:configuration_and_datapipeline}---is often preferable to creating entirely new components, as it encourages modularity and reuse. Since many of these components rely on initialization from structured data stored in files, Section~\ref{sec:storage_and_retrieval} describes initializer functions commonly used to populate component attributes introduced in the following subsections.

\subsubsection{3D primitives}
These are the fundamental building blocks in 3D rendering and modelling. Common primitives include \textbf{point clouds}, \textbf{line sets}, and \textbf{meshes}, all of which are defined using spatial points, but differ in how those points are connected and interpreted:

\begin{itemize}
    \item \textbf{Point clouds} represent discrete spatial samples without connectivity information.
    \item \textbf{Line sets} add connectivity metadata, defining which points are linked to form lines.
    \item \textbf{Meshes} extend line sets by specifying how lines (edges) form surfaces (faces).
\end{itemize}

In mesh terminology, vertices, edges, and faces correspond respectively to the points, lines, and surfaces, as described above.

\subsubsection{3D Gaussian splats}

As an alternative to conventional mesh-based representations, 3D Gaussian Splats offer a high-fidelity method for modelling realistic scenes \cite{Kerbl:3Dgaussians}. Unlike meshes, where geometry is formed through connected elements, each splat is an independent primitive defined by a Gaussian distribution:

\begin{itemize}
    \item The \textbf{mean} of the Gaussian encodes the spatial \textbf{position}.
    \item The \textbf{covariance matrix} captures both the \textbf{shape} and \textbf{orientation} of an ellipsoid, allowing the splat to deform into ellipsoids of varying anisotropy.
    \item \textbf{Color} and \textbf{opacity} describe visual appearance.
\end{itemize}

These ellipsoidal splats can effectively represent a wide range of geometry through its scale and orientation: for example, long thin splats approximate beams or poles, while large, flat splats represent transparent surfaces. In illustration on bottom of Figure \ref{fig:3dgs_from_images}, the hull of the boat consist of large flat orange splats, with darker thin splats representing screw on the boat.

Due to their point-based nature, Gaussian splats integrate naturally with PyGemini's architecture. Their spatial property can be stored in point clouds and visual properties are encapsulated as a separate splat component associated with each point, enabling flexible querying, filtering, and manipulation within the simulation pipeline.

\subsubsection{Rosbag} \label{sec:rosbag}
PyGemini supports both rosbag versions 1 and 2--ROS’s standardized log file format for recording message data, can be both read, written and edited. Because Rosbags are based on portable serialization and database formats, this integration does not require the ROS middleware or runtime environment. As such, it is ideal for offline workflows such as data analysis, post-mission evaluation, or augmentation of recorded sensor data. A typical use case includes image enhancement or in-painting (Figure~\ref{fig:sd_pipeline}), where minimal external dependencies are desirable and real-time interaction is not required.

\subsubsection{Trajectories}
Are a set of 2D points in a xy-coordinate frame, along with timestamps. The latter is usually in POSIX time, but can vary depending on needs. They are often interpolated to be synchronised with other trajectories of different timestamps or to reflect a reference time driven by e.g. real sensor data coming from rosbags.

\subsubsection{Storage and retrieval} \label{sec:storage_and_retrieval}
PyGemini can store and retrieve content from various sources and locations with use of initializer functions described in Section \ref{sec:library}. Several initialiser functions allow for asset retrieval such as mesh, splat, and vessel trajectories, in addition to datasets such as rosbags. This can happen locally, on a network drive or via external web addresses. For the latter, this paper contributes with content accessible at \cite{Vasstein:PyGemini_Data} that can be used by these initializer functions.
\section{Use cases} \label{sec:use_cases}
\begin{table*}[!ht]
    \centering
    \renewcommand{\arraystretch}{1.2}
    \setlength{\tabcolsep}{6pt}
    \begin{tabular}{l|c|c|c|c|c|c|c}
        \hline
        \textbf{} & \multicolumn{7}{c}{\textbf{Applications}} \\
        \hline
        \textbf{Processors} & \makecell{\textbf{Image} \\ \textbf{Rendering}} & \makecell{\textbf{Target Tracking} \\ \textbf{Simulation}} & \makecell{\textbf{Image} \\ \textbf{Augmentation}} & \makecell{\textbf{Interactive} \\ \textbf{Trajectory} \\ \textbf{Planning}} & \makecell{\textbf{3DGS} \\ \textbf{from Rosbags}} & \makecell{\textbf{3DGS} \\ \textbf{from Images}} & \makecell{\textbf{Image} \\ \textbf{Labeling}} \\
        \hline
        Viser                             & \cmark{Yes} & \cmark{Yes} & \cmark{Yes} & \cmark{Yes} & \cmark{No}  & \cmark{No}  & \cmark{Yes} \\ 
        Open3D                            & \cmark{Yes} & \cmark{Yes} & \cmark{Yes} & \cmark{Yes} & \cmark{No}  & \cmark{No}  & \cmark{Yes} \\ 
        Rosbag reader                     & \cmark{Yes} & \cmark{No}  & \cmark{Yes} & \cmark{No}  & \cmark{Yes} & \cmark{No}  & \cmark{Yes} \\
        Segmentation                      & \cmark{No}  & \cmark{No}  & \cmark{Yes} & \cmark{No}  & \cmark{Yes} & \cmark{Yes} & \cmark{No}  \\ 
        3DGS Trainer                      & \cmark{Yes} & \cmark{No}  & \cmark{No}  & \cmark{No}  & \cmark{Yes} & \cmark{Yes} & \cmark{No}  \\
        Trajectory reader                 & \cmark{Yes} & \cmark{Yes} & \cmark{No}  & \cmark{Yes} & \cmark{No}  & \cmark{No}  & \cmark{No}  \\
        Trajectory planner                & \cmark{Yes} & \cmark{Yes} & \cmark{No}  & \cmark{Yes} & \cmark{No}  & \cmark{No}  & \cmark{No}  \\
        Agent-based trajectory generation & \cmark{Yes} & \cmark{Yes} & \cmark{No}  & \cmark{Yes} & \cmark{No}  & \cmark{No}  & \cmark{No}  \\
        SeaState                          & \cmark{Yes} & \cmark{Yes} & \cmark{No}  & \cmark{No}  & \cmark{No}  & \cmark{No}  & \cmark{No}  \\
        Lidar simulation                  & \cmark{Yes} & \cmark{Yes} & \cmark{No}  & \cmark{No}  & \cmark{No}  & \cmark{No}  & \cmark{No}  \\
        Structure from Motion             & \cmark{No}  & \cmark{No}  & \cmark{No}  & \cmark{No}  & \cmark{Yes} & \cmark{Yes} & \cmark{No}  \\ 
        Target Tracker                    & \cmark{No}  & \cmark{Yes} & \cmark{No}  & \cmark{No}  & \cmark{No}  & \cmark{No}  & \cmark{No}  \\
        Real-Image Inpainting             & \cmark{No}  & \cmark{No}  & \cmark{Yes} & \cmark{No}  & \cmark{No}  & \cmark{No}  & \cmark{No}  \\
        ControlNet Rendering              & \cmark{No}  & \cmark{No}  & \cmark{Yes} & \cmark{No}  & \cmark{No}  & \cmark{No}  & \cmark{No}  \\
        Gaussian Splat Enhancement        & \cmark{No}  & \cmark{No}  & \cmark{Yes} & \cmark{No}  & \cmark{No}  & \cmark{No}  & \cmark{No}  \\
        \hline
    \end{tabular}
    \newline
    \caption{Processor usage across PyGemini applications. Columns represent key applications; rows represent individual processors. A check mark indicates usage within that application.}
    \label{tab:processor_usage}
\end{table*}

Several use cases have been developed to demonstrate how PyGemini can support a range of domains relevant to maritime autonomy. These use cases are implemented as software applications, where each application consists of components and processors described within the context of their primary function. Table~\ref{tab:processor_usage} provides an overview of processor reuse across applications, highlighting maintainability benefits of the ECS architecture. In addition, Figure~\ref{fig:service_diagram} presents a higher-level diagram illustrating how these applications may interact when deployed as cloud-based services.

\subsection{3DGS from Images}

\begin{figure}[!ht]
    \centering
    \includegraphics[width=0.8\linewidth]{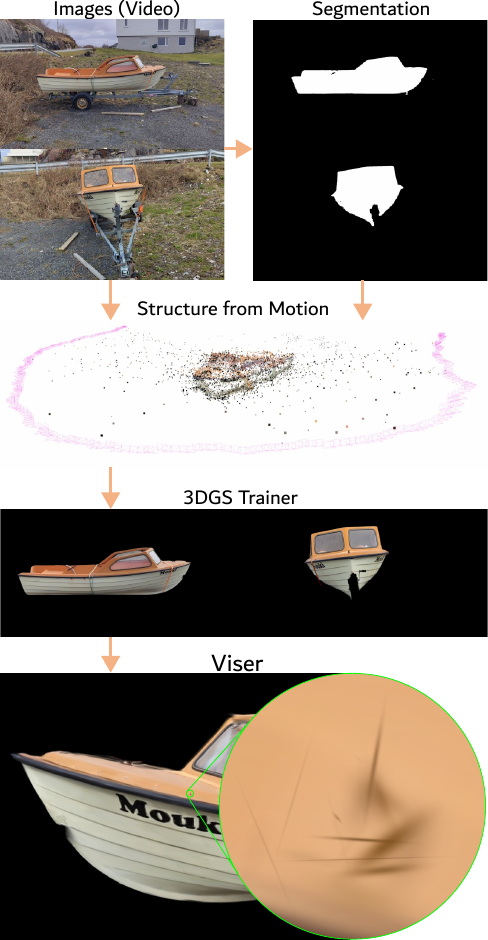}
    \caption{Overview of processes involved in creating a 3DGS model from images sampled from videos.}
    \label{fig:3dgs_from_images}
\end{figure}

As illustrated in Figure \ref{fig:service_diagram}, 3D models are produced through dedicated services related to 3D Gaussian Splatting (3DGS). For the case of 3DGS from images, the service consists of segmentation, Structure from Motion (SfM) and 3DGS trainer. In this use case, images are sampled from videos, with camera transforms and intrinsics generated from the SfM process. The 3DGS trainer then trains based on component inputs and eventually yields a 3DGS model, which is represented by components for point cloud, and splats. Example vessel models are shown in Figure \ref{fig:3dgs_models}, while the model generation process is illustrated in Figure \ref{fig:3dgs_from_images}.

\subsubsection{Segmentation}
In various applications, isolating regions of interest (ROI) from disturbances is essential. Disturbances could be unwanted semantics such as moving vehicles, visible camera housing, or rain droplets stuck to camera lens. These are unwanted elements that could distort the final 3D model. Binary masks are therefore used to indicate which parts of an image are of interest. PyGemini integrates these via segmentation models, that defines the regions to be included or excluded. These masks can subsequently be utilized by downstream modules for data filtering, or to provide additional training context.

\subsubsection{Structure from Motion}
To enable 3D reconstruction from images, both intrinsic and extrinsic camera parameters must be estimated. PyGemini incorporates a SfM processor that estimates these parameters, and a sparse 3D point cloud from input images. Parameters can also be initialized to guide the optimization closer to the ground truth---accelerating convergence, and improving robustness in ambiguous scenes \cite{schoenberger:sfm}.

\subsubsection{3DGS Trainer}
The 3DGS training pipeline uses the component interface in Section \ref{sec:components} to compose models from various sources of inputs---involving images, camera intrinsics, poses, and sparse 3D points. These elements are handled as modular components associated with entities. In this context, each camera is represented as an entity with corresponding image, pose, and intrinsic parameters across various components. This design decouples the original data format from the internal representation used by processors via the component interface. This enables a flexible and reusable input mechanism across use cases.

The training output gives a point cloud component where each point represents the mean position of a Gaussian splat. Associated attributes---such as color, opacity, and covariance---are stored separately in a dedicated splat component. This allows results to either be visualized as a point cloud or 3DGS model, that is further discussed in Section (\ref{sec:image_rendering}).

\begin{figure}[!ht]
    \centering
    \includegraphics[width=0.32\linewidth]{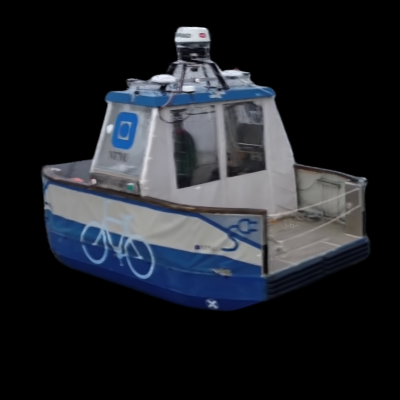}
    \includegraphics[width=0.32\linewidth]{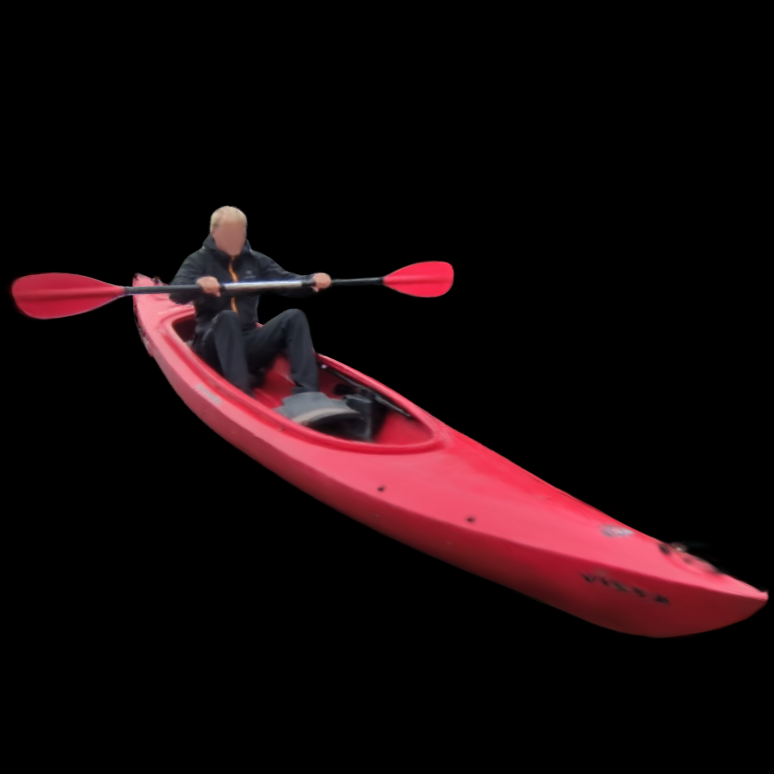}
    \includegraphics[width=0.32\linewidth]{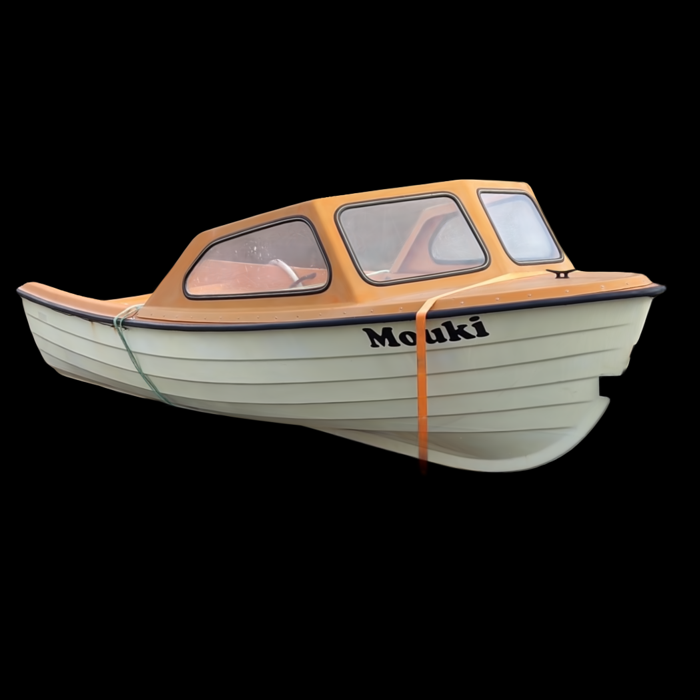}
    \includegraphics[width=0.32\linewidth]{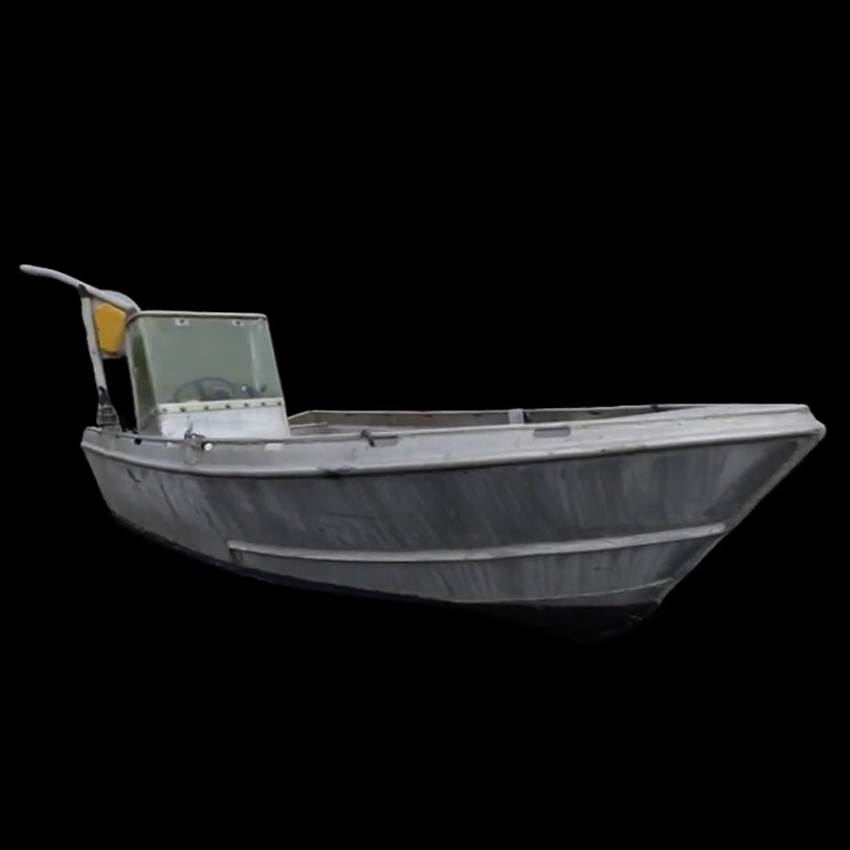}
    \includegraphics[width=0.32\linewidth]{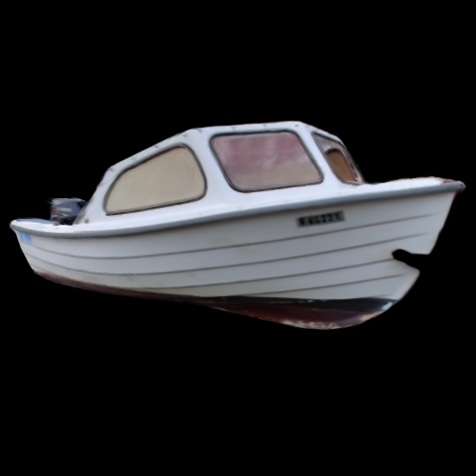}
    \includegraphics[width=0.32\linewidth]{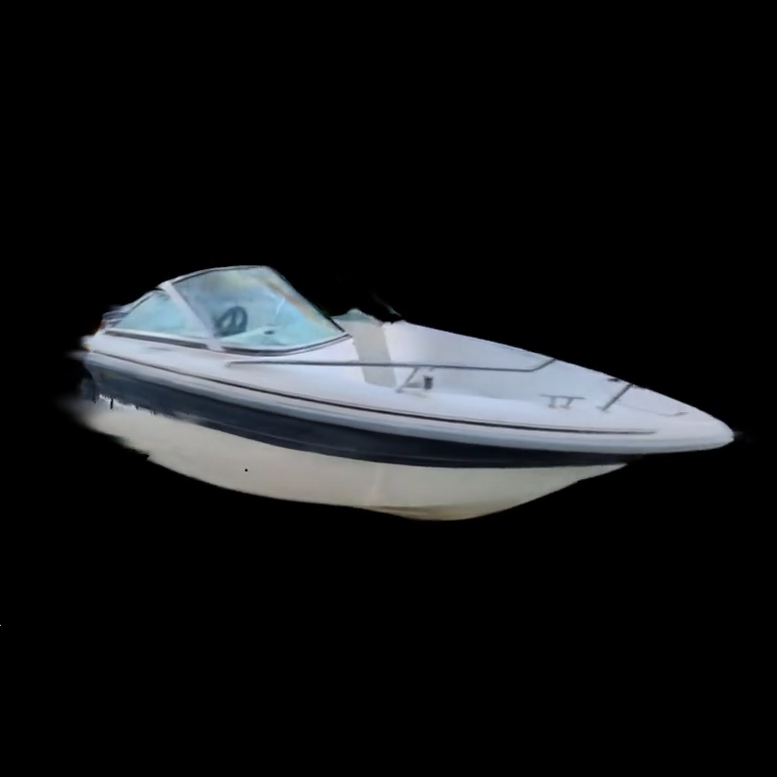}
    \caption{Generated 3DGS models of maritime vessels.}
    \label{fig:3dgs_models}
\end{figure}

\subsection{3DGS from Rosbags}

\begin{figure}[!ht]
    \centering
    \includegraphics[width=\linewidth]{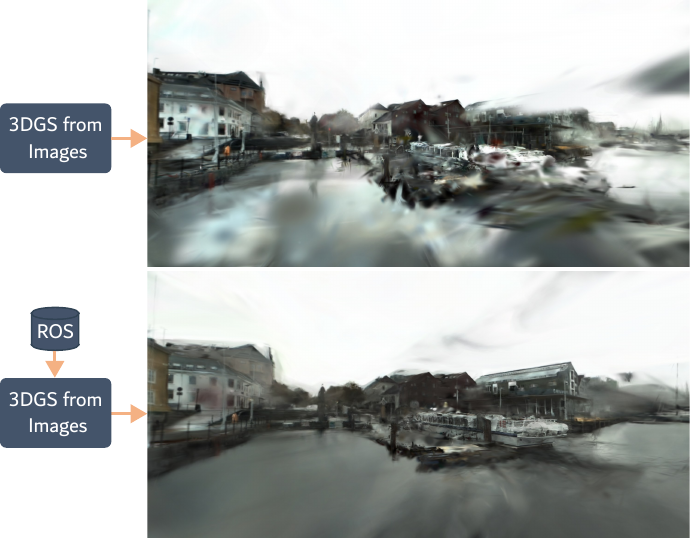}
    \caption{Overview of "3DGS from images" service (upper) and it's improvement through use of additional rosbag data (lower). Details of "3DGS from images" service is shown in Figure \ref{fig:3dgs_from_images}.}
    \label{fig:3dgs_from_rosbags}
\end{figure}

Instead of initializing solely from images, PyGemini can leverage additional data sources, such as lidar and navigation data from rosbags. This service utilizes multimodal sensor inputs to generate more accurate initial estimates for camera poses and lidar-based point clouds.
Having rough estimates of camera positions and where dense objects are located in the scene helps prevent errors in camera pose estimation and 3D point reconstruction. These advantages prove especially useful in uncontrolled environments, where images alone are prone to disturbances. With estimation being based on multiple sensors, disturbances and inaccuracies are better handled. Resulting reconstructions achieves higher-quality compared to the image-only pipeline, given that the measurements are sufficiently accurate and refined in SfM or similar methods. Additional sensor data from the rosbag is also observed to significantly improve SfM performance. In certain cases, the percentage of registered images used in the reconstruction increases from 18 to 81\%. This can be seen in Figure \ref{fig:3dgs_from_rosbags}, where the two 3DGS services are compared.

\subsubsection{Rosbag reader} \label{sec:ROS_reader}
Rosbag topics can be directly converted to component data from several specialized processors. This involves large data formats such as image data from various camera formats, point clouds from lidar and radar, in addition to time series data such as GNSS position, heading, etc. 

\subsection{Interactive Trajectory Planning}
\begin{figure}[!ht]
    \centering
    \includegraphics[width=0.9 \linewidth]{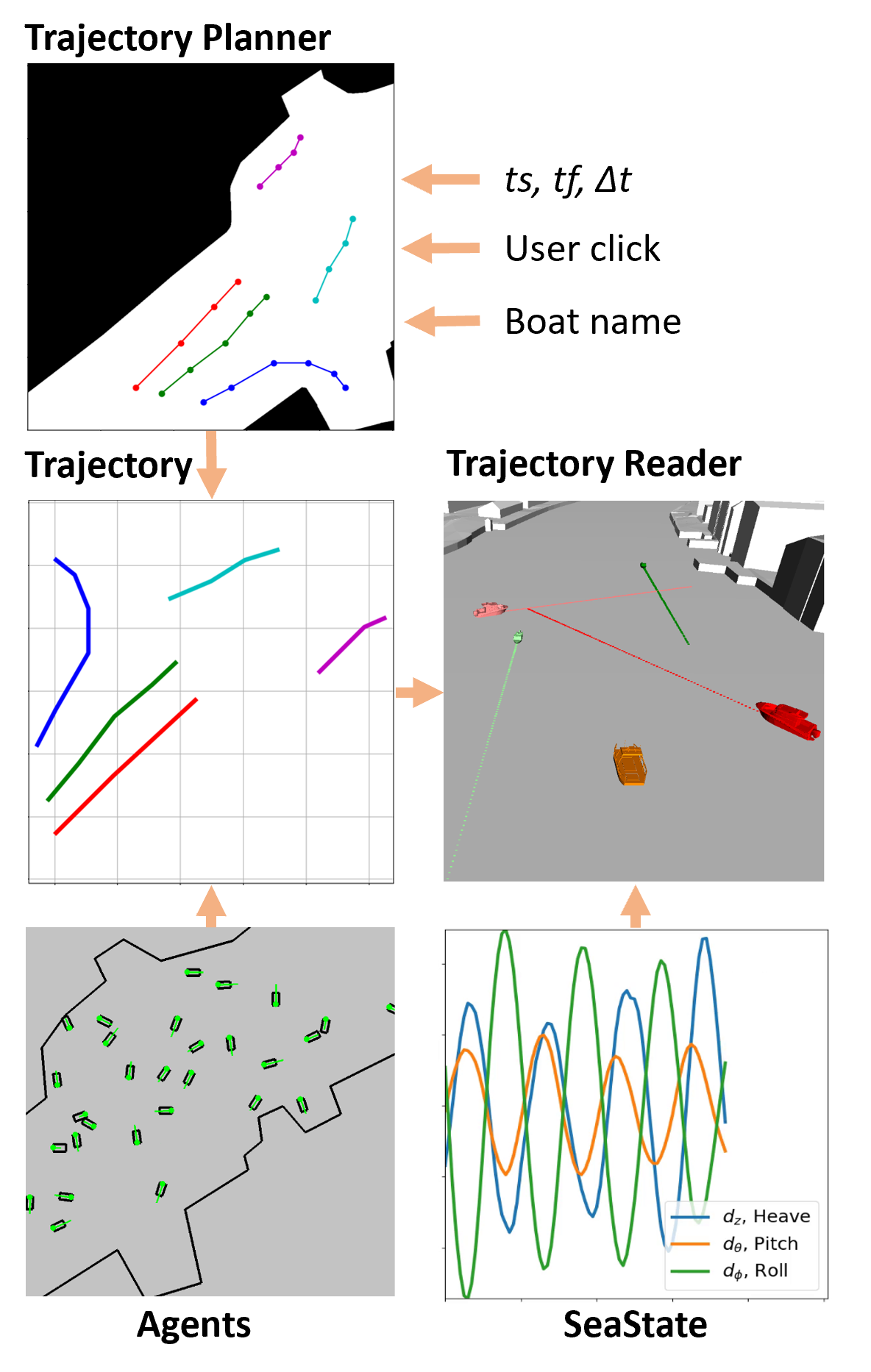}
    \caption{An overview of the Scenario Generation module. The Trajectories might be created by the clickable Trajectory planner or external tools, such as Agents. The resulting trajectories are read from the file and dictate the motion of the objects. Additionally, SeaState can simulate wave effects on floating objects.}
    \label{fig:ScenarioGeneration}
\end{figure}

The Interactive Trajectory Planning application, shown in Figure~\ref{fig:ScenarioGeneration}, focuses on simulating the motion of target objects. Twenty scenarios were created to investigate the effects of occlusions. The majority of trajectories were generated using the Trajectory Planner tool, which allows users to define paths by interacting with a 2D image interface.

\subsubsection{Trajectory planner}
The Trajectory planner produces trajectory files dictating the targets' three degrees of freedom (DOF) position and heading at various timestamps. The user places waypoints for each boat using a clickable 2D view of the harbour. These waypoints can be modified at a later stage. The tool then transforms the waypoints into world coordinates and interpolates between them using constant-velocity motion to generate smooth trajectories, which are written to a file. 

\subsubsection{Agent-based trajectory generation}
External tools can be used to create trajectories. One example involves agent-based control of targets, as demonstrated by adapting the code from \cite{Fjellheim:Agents}. The user specifies the free space, agent type, and starting positions, after which trajectories are automatically generated. These can then be imported into PyGemini.

\subsubsection{Trajectory reader}
The trajectory reader is the processor responsible for correctly placing objects in the world, given the specification in the trajectories. These trajectories can come from processors such as "Agent-based trajectory generation" or "Trajectory planner". Alternatively, pre-recorded and preprocessed trajectories from real-world data, or external sources, may also be loaded directly using initializer functions (Section \ref{sec:storage_and_retrieval}).

\subsection{Target Tracking Simulation}
\begin{figure}[!ht]
    \centering
    \includegraphics[width=0.8\linewidth]{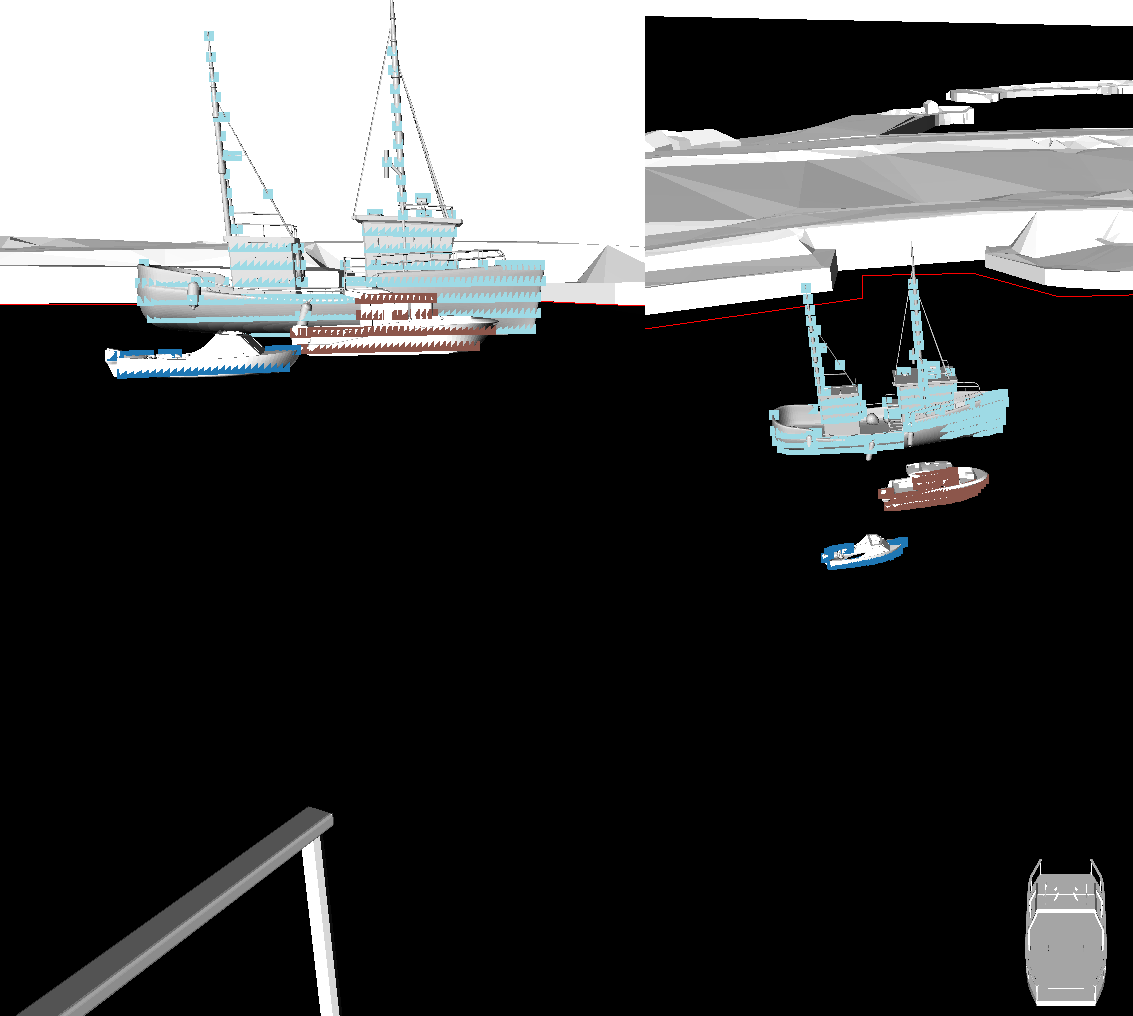}
    \caption{Target tracking scene in Open3D making use of the lidar measurements.}
    \label{fig:tracking_example}
\end{figure}
Target tracking simulations consist of a target tracker, lidar simulation, the scenario generation, and the camera visualizer. The scenarios can be used to validate trackers quantitatively using performance metrics. Figure~\ref{fig:tracking_example} illustrates an example target tracking scene, visualized with a lidar-generated point cloud.

\subsubsection{Target Tracker}
Target tracking involves associating sensor measurements with targets and updating state estimates accordingly. In PyGemini, the VIMMJIPDA tracker~\cite{Brekke:VIMMJIPDA} has been integrated to perform this task.

\subsubsection{SeaState}
SeaState is included to simulate the motion of floating objects, providing realistic movement for sensors such as cameras and lidar. Currently, two models are available. The first applies a deterministic sine-based disturbance to each object's orientation. The second introduces a second-order autoregressive disturbance model that stochastically affects both the orientation and vertical position of the entity relative to the water surface.    

\subsubsection{Lidar simulation} \label{sec:Lidar_simulation}
The lidar simulator uses CPU based ray tracing from Open3D \cite{Zhou:open_3d}, where a functionality for ray drop is added based on simulated intensity levels. See Appendix \ref{app:lidar_intensity} for further details.

\subsection{Image Rendering} \label{sec:image_rendering}
\begin{figure}
    \centering
    \includegraphics[width=0.9\linewidth]{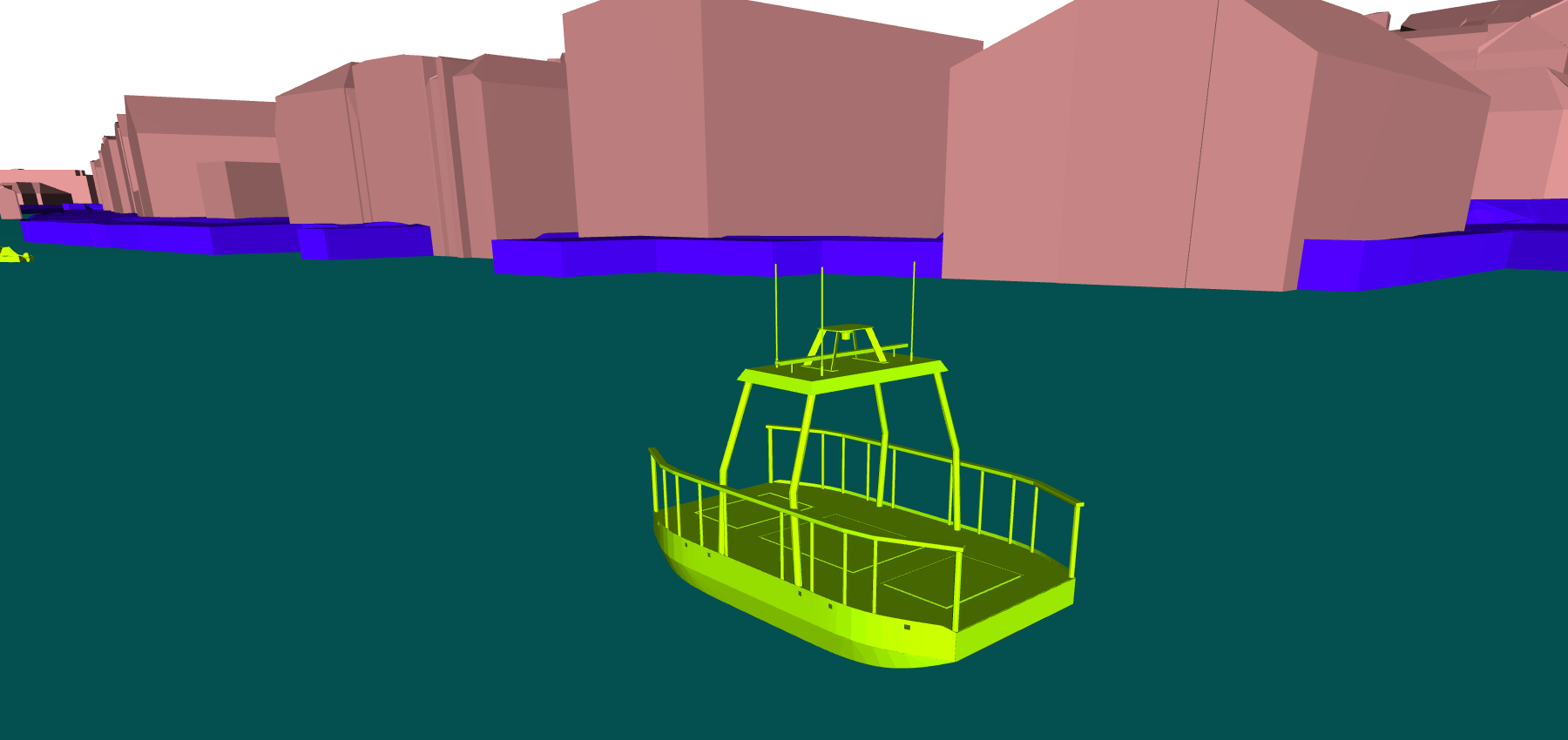}
    \includegraphics[width=0.9\linewidth]{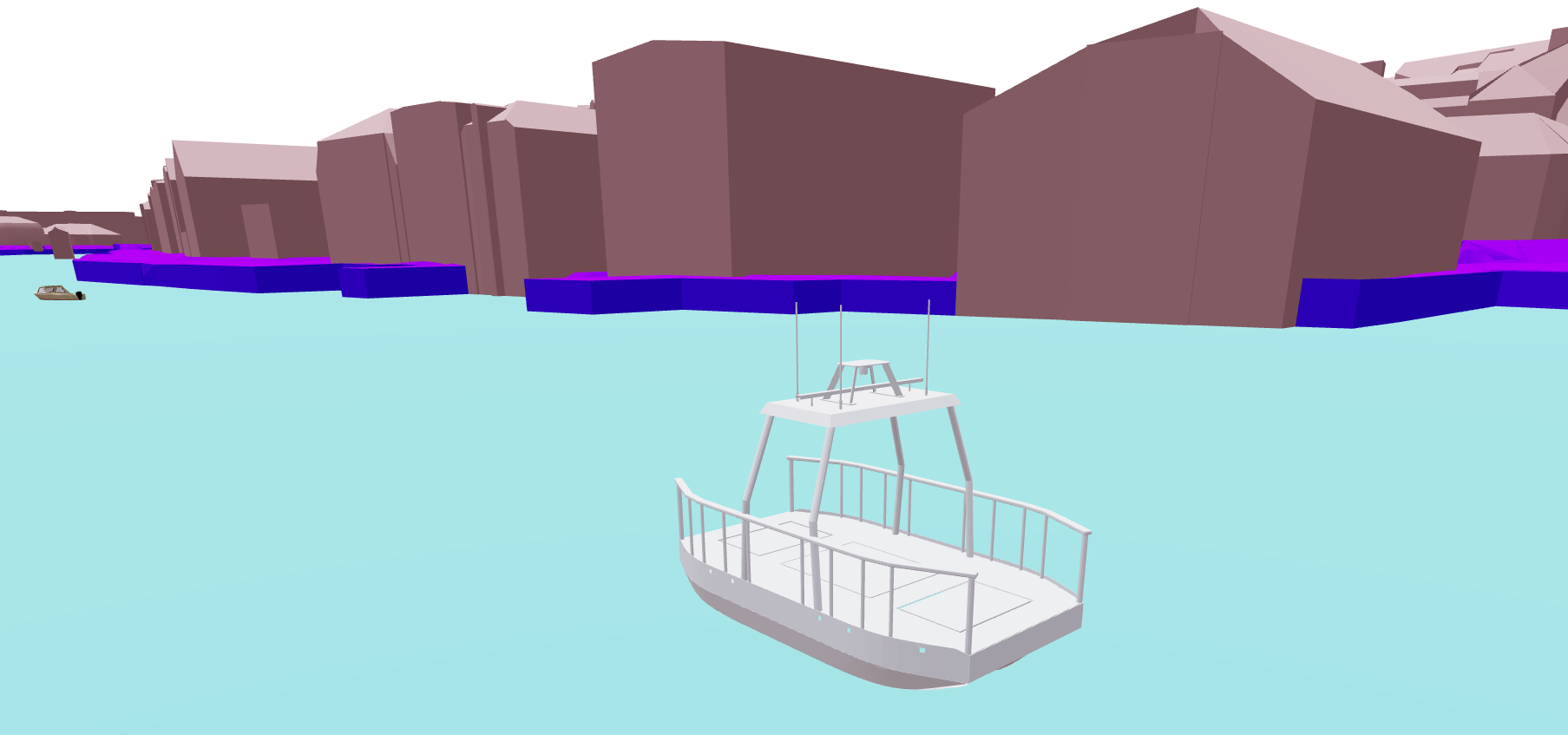}
    \includegraphics[width=0.9\linewidth]{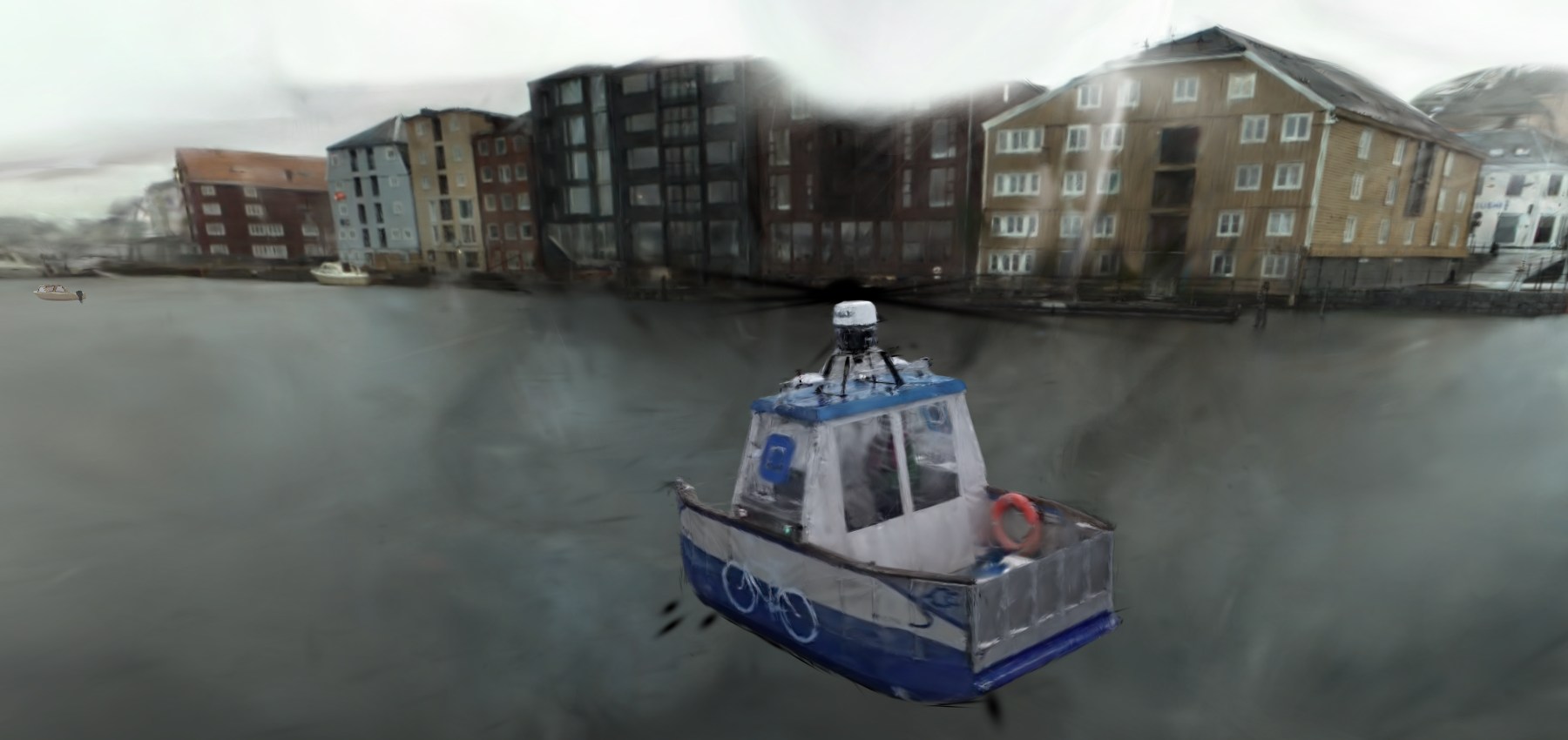}
    \caption{Top to bottom: A scenario visualized in various formats and visualizers. First two are mesh based, presented in Open3D and Viser. The last uses 3DGS instead of mesh in Viser.}
    \label{fig:scenario_visualization}
\end{figure}

PyGemini supports two camera rendering systems: Open3D \cite{Zhou:open_3d} and Viser from Nerfstudio \cite{Tancik:nerfstudio}. This allows for multiple capabilities for rendering 3D primitives and 3DGS in different contexts. Open3D is mainly used in legacy systems and specific image augmentations (such as in Section \ref{sec:controlnet_rendering}), while Viser allows for more flexible, scene-aware inspection. Viser also supports rendering 3DGS models from arbitrary camera positions, generally resulting in higher fidelity images than Open3D (Figure \ref{fig:scenario_visualization}). Here, 3DGS models can be rendered in conjunction with other 3D primitives, such as meshes, point clouds, and linesets.

\subsubsection{Open3D}
Open3D is an open-source library with general purpose support for 3D data operation and visualization \cite{Zhou:open_3d}. In PyGemini, Open3D serves both as component representation for standard in-library 3D primitives and viewer. The state of the 3D primitives components, such as meshes and 3DGS, can be visualized in its 3D renderer and viewer. In addition to providing rendering it also provides depth, segmentation and normal maps from the scene as shown in Figure \ref{fig:sd_pipeline}.

\subsubsection{Viser}
Viser is developed as part of the Nerfstudio project \cite{Tancik:nerfstudio}, providing both visualisation and systems for designing GUIs. The project itself focuses on various Novel View Sythesis (NVS) techniques, and in particular supports native rendering of 3DGS. 3D primitives are mainly represented using primitive data types, allowing for visualizing of arbitrary data. This can be retrieved from Open3D primitives, thus can be used interchangeably with Open3D (as shown in Figure \ref{fig:scenario_visualization}). Also, unlike its counterpart, it is a web-based renderer with a server-client architecture, where rendering is performed on client side. For instance, back-end processes run on the server machine, while a client node can operate either on the same machine or on a remote one. This provides a flexible option for choosing where the front-end rendering takes place.

\subsubsection{ROS Node} \label{sec:ROS_node}
PyGemini can through the rclpy library from ROS operate as a native ROS 2 node. A docker container has been made to handle software dependencies, enabling real-time interaction with live systems through standard ROS topics, services, and actions. Several dedicated processors can publish, subscribe, and participate directly in ROS communication graphs. This setup can be used for real-time testing---including HIL and SIL---or providing visualisation systems for autonomy. 

\subsection{Image Labeling}
 \begin{figure}[!ht]
    \centering
    \includegraphics[width=1\linewidth]{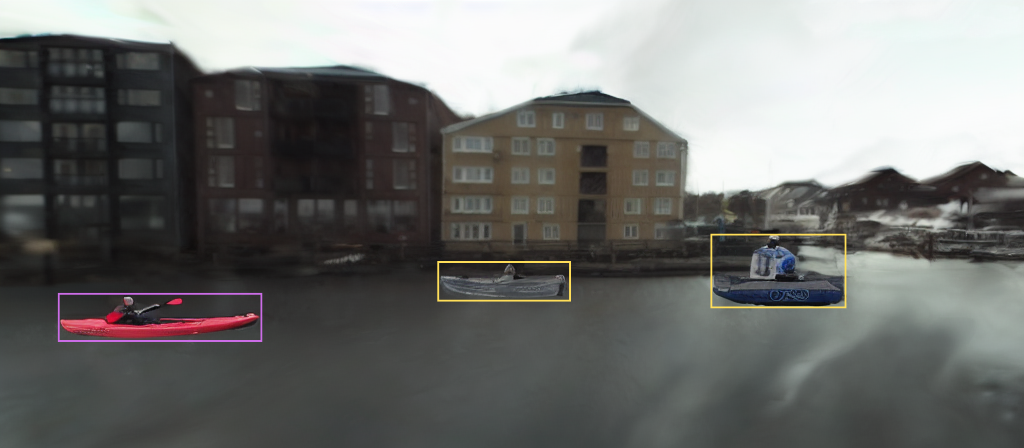}
    \caption{Automatic bounding box generation for target boats from a 3DGS enhanced image (Section \ref{sec:3dgs_enhancement}). The targets may also be categorized into different classes, such as boats and kayaks.}
    \label{fig:BoundingBox_viser}
\end{figure}
 Knowing locations and extents precisely enables the automation of processes, such as bounding box annotation. These annotations can be exported to a file and used to train object detectors. Figure \ref{fig:BoundingBox_viser} presents the automatically annotated bounding boxes in the Viser visualizer.

\subsubsection{Bounding Box generation}
2D Bounding Boxes can be automatically computed, rendered, and exported. This enables downstream tasks such as validating and training visual object detectors that rely on accurate localization information.

\subsection{Image Augmentation}
\begin{figure}[ht]
    \centering
    \includegraphics[width=\linewidth]{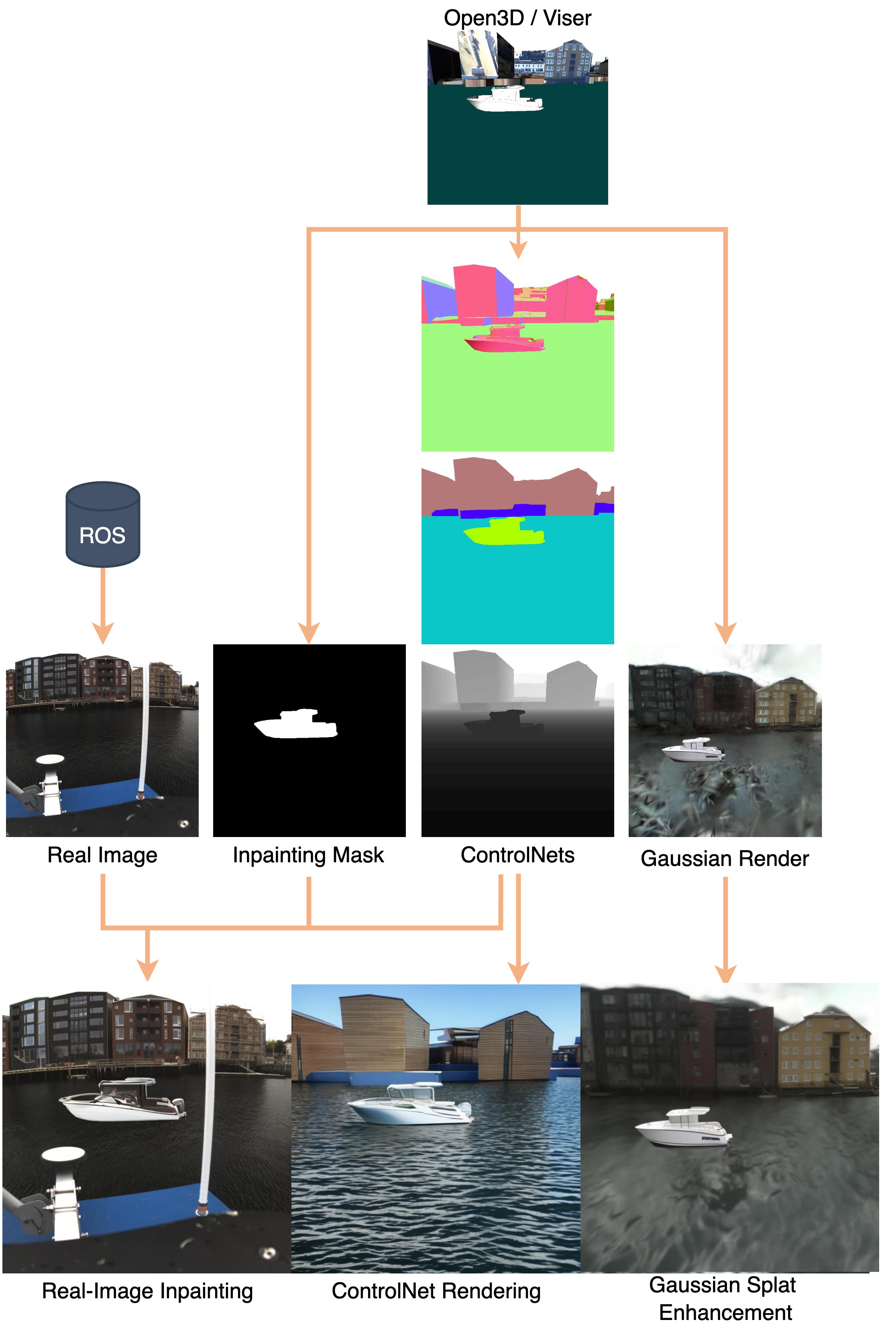}
    \caption{Overview of the Stable Diffusion pipelines in PyGemini.}
    \label{fig:sd_pipeline}
\end{figure}

To generate augmented images, PyGemini integrates Stable Diffusion (SD), an artificial intelligence-based image generator. It creates images by gradually refining random noise through an iterative denoising process \cite{Rombach:Latent_diffusion}. The generation can be guided by prompts or ControlNets, enabling both text-based and spatial control over the output \cite{zhang:ControlNets}. In PyGemini, this functionality is implemented in three different ways through dedicated processors, one that handles inpainting, simulator-driven image generation and enhancement of Gaussian Splat renderings, as illustrated in Figure~\ref{fig:sd_pipeline}. Prompts are defined in a configuration file and passed into the pipeline through a modular SD component.

\subsubsection{ControlNet Rendering} \label{sec:controlnet_rendering}
This processor uses simulator-generated ControlNet inputs to produce synthetic maritime images. Open3D is used to generate depth, normal, and segmentation maps from 3D scenes, which guide the image generation process. By conditioning on these spatial inputs, SD maintains correct object placement, scale, and surface orientation. The resulting image, shown in the center branch of Figure \ref{fig:sd_pipeline}, feature a synthetic boat rendered without relying on detailed meshes or textures.

\subsubsection{Real-Image Inpainting}
The hybrid pipeline uses real images from ROS and augments them through Inpainting. The Open3D renderer generates ControlNet images along with binary masks that define the regions where new content should be inserted. Any 3D model can be rendered in the simulator to produce these inputs, enabling diversity in shape and appearance. Together, the ControlNet images and mask guide SD to generate a new boat within the editable area. This enables insertion of synthetic maritime objects with full background realism, allowing generation of a wide variety of boat renderings in authentic scenes. This process is illustrated in the left branch of Figure \ref{fig:sd_pipeline}.

\subsubsection{Gaussian Splat Enhancement} \label{sec:3dgs_enhancement}
The right branch of Figure \ref{fig:sd_pipeline} shows the pipeline used to enhance images generated through Gaussian splatting. It is also responsible for the teaser image at the first page of this paper. Although splatting offers an efficient approach to 3D reconstruction, the resulting images often include visual noise and lack fine detail. This pipeline applies SD Image-to-image \cite{Rombach:Latent_diffusion} to reduce noise and produce sharper image content.

\subsection{Applications as services}
\begin{figure}[!ht]
    \centering
    \includegraphics[width=0.9\linewidth]{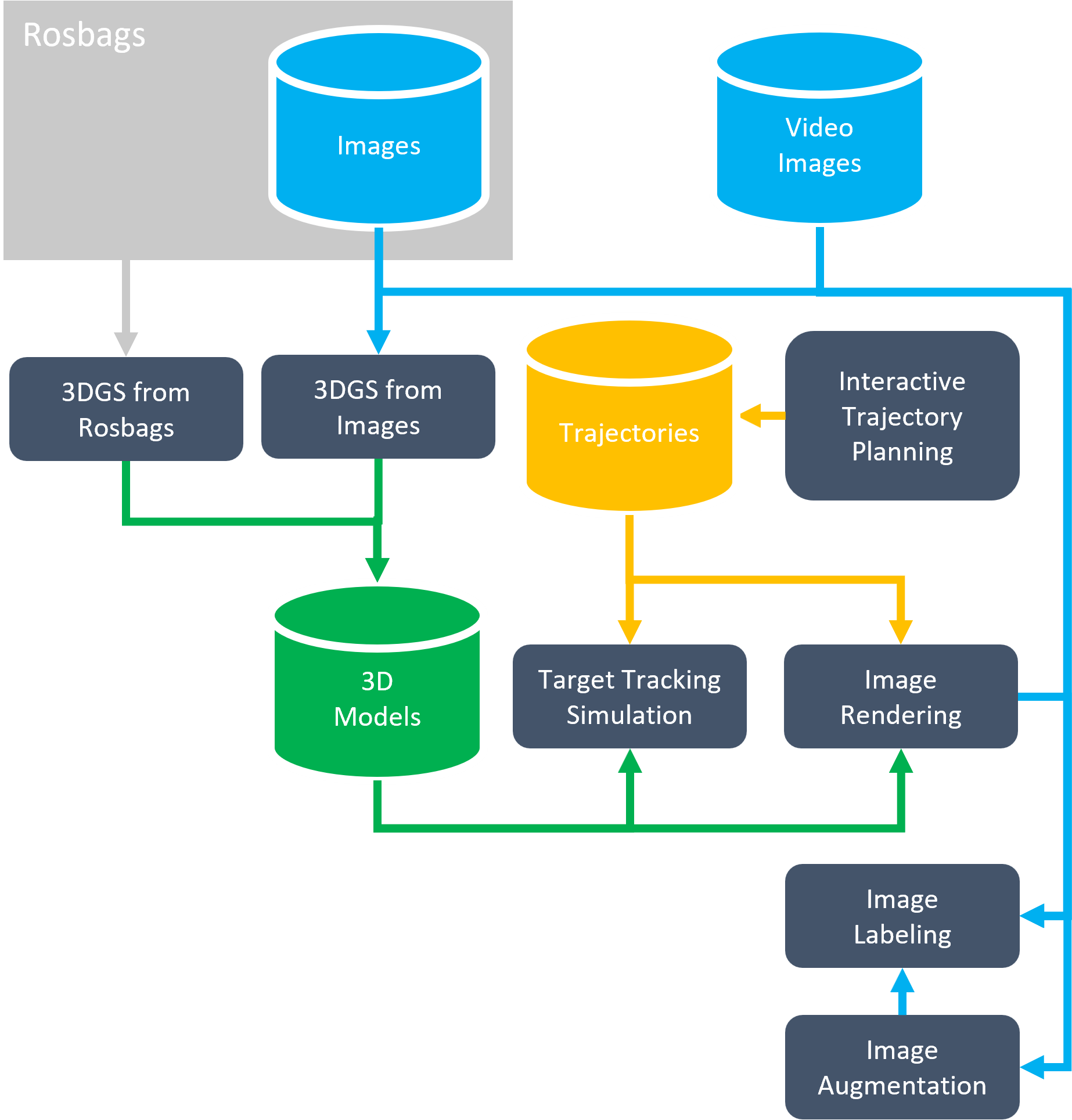}
    \caption{Higher-order data pipelines showcasing how the applications can influence each other if deployed as services, e.g. in cloud systems. Processor composition for each application is described in Table \ref{tab:processor_usage}.}
    \label{fig:service_diagram}
\end{figure}
Following the individual presentation of each application, we now describe how PyGemini supports scalable deployment by running applications as independent services that can communicate and exchange data within distributed computing environments (Figure \ref{fig:service_diagram}).

Through its integration with Docker \cite{Merkel:Docker} and a modular interface built on components (Section~\ref{sec:components}), PyGemini supports both on-premises and cloud-based deployment. This flexibility allows resource-intensive tasks---such as image augmentation or 3DGS model generation---to be offloaded to dedicated services, reducing bottlenecks and enabling continuous data delivery to other services such as simulators or image labeling pipelines. 

Furthermore, databases and file servers can be interconnected via shared data streams, enabling efficient exchange of resources such as 3D models, images from various data sources, and trajectory data, as illustrated in Figure~\ref{fig:service_diagram}. These services are not limited to PyGemini-based systems. For instance, several resources available from PyGemini's data repository \cite{Vasstein:PyGemini_Data}, allows generated 3D models to be shared with external systems, such as engine extensions \cite{Vasstein_autoferry_gemini, Loncar:Marus, Potokar:holoocean}. This allows any compatible service, regardless of its underlying framework, to benefit from data produced within the PyGemini ecosystem.

\section{Discussion}
In this section, we step back from the detailed presentation of PyGemini’s architecture and applications to critically interpret our empirical findings in the broader context of autonomous maritime systems, illustrated in Table \ref{tab:feature_comparison}. We close the loop on current challenges we face to transparently acknowledge our remaining gaps, and sketch a clear agenda for further research.

\subsection{Communication}
A key differentiator of PyGemini is its flexible and Python-native support for both message-passing and shared-memory communication, bridging capabilities found in existing solutions like ROS, OSP, and game engines.

Unlike ROS and OSP, which rely exclusively on message-passing architectures---ROS through anonymous publish-subscribe topics and synchronous client-server services, and OSP using FMI-based co-simulation buses---PyGemini supports message-passing as a first-class citizen while also enabling direct GPU–GPU and shared-memory data exchanges for high-performance pipelines. Through native Python bindings, PyGemini seamlessly integrates with ROS nodes (see Section~\ref{sec:ROS_node}), similar to HoloOcean and Stonefish, but with a more extensible architecture that aligns with Python-based autonomy R\&D workflows.

In contrast, platforms like Stonefish, HoloOcean, and Autoferry Gemini are tightly coupled to their respective simulation engines, often relying on custom network bridges or rigid engine APIs for inter-process communication. While these tools provide GPU-accelerated sensor models---Stonefish for underwater vehicles, Autoferry Gemini for ASVs---they lack a unified API for leveraging this across modular research pipelines. Additionally, integration with broader co-simulation tools like OSP remains limited to Autoferry Gemini and Vico.

PyGemini advances this ecosystem by offering out-of-the-box support for GPU–GPU data pipelines through its Python ecosystem. While it currently does not feature native GPU-accelerated sensor models, it builds upon libraries that enable high-performance workflows---such as running diffusion models or training neural radiance fields, as demonstrated in Figures~\ref{fig:sd_pipeline} and~\ref{fig:3dgs_from_images}. Such capabilities are difficult to replicate for the other contenders, due to their lack of support for modern machine learning and visualization libraries, which PyGemini leverages extensively through Python.

\subsection{Simulation}
Unlike traditional platforms such as ROS and OSP, which depend on external simulators, PyGemini offers a more modular and extensible approach to simulation tailored for autonomous maritime systems. While it currently does not support full motion physics (focusing instead on kinematics), it is uniquely designed to enable sensor-centric simulations that can be flexibly composed and interchanged.

The latter allows simulation fidelity to be modulated, by swapping processors while preserving underlying component data. This allows consistent scene representation across different sensor models, as illustrated in Figure~\ref{fig:scenario_visualization}, making it easier to explore trade-offs between performance and realism. While engine extensions based of Unity and Unreal allow for some of the same fidelity scaling, it comes at the cost of a less flexible and tightly coupled solution that halters portability.

In terms of sensor support, PyGemini matches or exceeds many existing tools:
\begin{itemize}
    \item It supports both RGB and thermal cameras through 3DGS, exceeding the rendering capabilities of engines such as Stonefish, while rivalling game engine extensions like Autoferry Gemini. In addition it supports both web based camera renderings (via Viser) and local rendering (via Open3D).
    \item Through its integration with neural rendering pipelines such as 3DGS, PyGemini can produce photorealistic content and supports training from various data sources (e.g., ROS bags, images, and videos), as seen in Figures~\ref{fig:3dgs_models} and \ref{fig:3dgs_from_rosbags}. These tools enable it to complement high-fidelity simulation engines by generating rich synthetic content, stored in a shared data repository \cite{Vasstein:PyGemini_Data}. With applications like 3DGS from sampled images from videos, it can potentially democratise 3D content---which until now, have been dominated by licenced materials. 
    \item On lidar, PyGemini and Autoferry Gemini are the only tools with in-built support for lidar simulation (Table \ref{tab:feature_comparison}). However, PyGemini goes further by simulating beam intensity variations based on surface properties, a feature that is only supported for radar in Autoferry Gemini.
\end{itemize}

Meanwhile, underwater-oriented platforms like Stonefish, HoloOcean, and MARUS focus heavily on sonar simulation. Stonefish includes advanced forward-looking sonar models; HoloOcean provides imaging, sidescan, and multibeam sonar; and MARUS offers a basic sonar system. These are not replicated in PyGemini, which is currently focused on surface vessels and above-water perception.

In summary, PyGemini’s simulation design emphasizes modularity, extensibility, and integration with modern ML-based rendering tools. While it does not yet support full physics dynamics, it provides a flexible and interoperable foundation for sensor-centric and data-driven simulation workflows, bridging gaps in existing simulators.

\subsection{Development}
Unlike many simulation platforms that are tightly coupled to specific engines or ecosystems, PyGemini is designed from the ground up for portability, extensibility, and integration into modern research workflows. Its architecture separates a lightweight, Python-native core library from optional GUI- and engine-based interfaces, making it uniquely suited for both framework-level execution and library-style integration.

Most competing tools offer some degree of cross-platform compatibility, often via containerization. However, engine-bound tools like MARUS, HoloOcean, and Autoferry Gemini are constrained by their dependence on Unity or Unreal, limiting reuse of core functionalities outside those engines. In contrast, large portions of PyGemini---as detailed in Sections~\ref{sec:library} and~\ref{sec:ECS}---are engine-agnostic and can be run natively in any Python-capable environment. This includes headless Docker deployments, ROS integration (Section~\ref{sec:ROS_node}), and direct interaction with sensor data through ROS bags (Section~\ref{sec:rosbag}).

Only a few tools---most notably ROS, Stonefish, and OSP---offer comparable flexibility through modular Python APIs (Table \ref{tab:feature_comparison}). However, PyGemini stands out as one of the only tools with a fully permissive, engine-independent license, offering complete source access and compatibility with standard Python tooling. While MARUS and HoloOcean offer permissive licensing, they remain constrained by their engines’ proprietary terms and internal APIs.

More significantly, PyGemini introduces a novel Configuration-Driven Development (CDD) workflow that addresses a major gap in existing tools: the lack of systematic, maintainable development processes. Where most alternatives rely on ad hoc configuration or custom scripting, PyGemini formalizes its development process through five core principles:
\begin{enumerate}
    \item \textbf{Shared Core Library}: The PyGemini core (Section~\ref{sec:library}) encapsulates shared logic that powers simulators, autonomy modules, and control interfaces alike. This promotes code reuse, reduces duplication, and ensures consistency across processors and applications.
    \item \textbf{Reusable Processors in ECS Architecture}: PyGemini’s ECS-based design supports pluggable, domain-agnostic processors (Table~\ref{tab:processor_usage}). This enables modular feature construction, allowing researchers to rapidly prototype new behaviours without disrupting existing logic.
    \item \textbf{Composable Configuration Files}: As seen in Figure~\ref{fig:configuration_and_datapipeline}, behaviours and system layouts are defined declaratively through structured configuration files, that also can be combined. These files bridge the gap between user intent and executable logic, forming the backbone of the CDD approach.
    \item  \textbf{Cloud-Agnostic Deployments}: The system’s deployment layer (Figure~\ref{fig:service_diagram}) is driven entirely by configuration, enabling seamless transitions between local, remote, or containerized environments. This allows the same codebase to scale across research labs, shore control centers, or cloud-hosted training clusters.
    \item \textbf{Automated Test Generation from Configurations}: PyGemini can automatically derive test cases from its configuration files (Table~\ref{tab:application_state}), aligning validation logic with user-defined system behaviour. This promotes maintainable regression testing and continuous integration without duplicating test logic.
\end{enumerate}
In essence, PyGemini transforms the way maritime autonomy tools are developed---from tightly coupled, manually maintained systems into modular, testable, and configuration-driven platforms. Its balance of permissiveness, integration with Python’s scientific ecosystem, and emphasis on long-term maintainability offers a development experience unmatched by current alternatives (Table \ref{tab:feature_comparison}).

\subsection{Challenges and Limitations}
Despite its modular design, PyGemini faces several adoption and technical hurdles. While some are addressable through tooling or design extensions, others stem from inherent language and architecture choices.

\subsubsection{User–Developer Transition}  
CDD lowers barriers by allowing users to work via configuration files. However, implementing custom features---like advanced rendering or new sensors---requires developer-level skills. This can limit contributions if the user base lacks programming expertise. Expanding the library of reusable processors and leveraging third-party Python packages can help bridge this gap.

\subsubsection{Real-Time Limitations}  
Python’s dynamic execution, garbage collection, and global interpreter lock introduce runtime unpredictability, making it unsuitable for hard real-time systems. PyGemini is best suited as an R\&D and simulation framework, not as a runtime for mission-critical onboard autonomy.

\subsubsection{Hardware and GPU Constraints}  
While containerization ensures reproducibility, it cannot bypass hardware limitations. GPU-intensive tasks may fail on weaker systems. However, PyGemini’s ECS model supports partial execution (e.g., lidar without visualization), enabling fallback modes that are rare in other solutions.

\subsubsection{Configuration Complexity}  
As systems scale, configuration files grow harder to manage. Without editor support or schema validation, errors may increase. Introducing linting, autocompletion, and schema-aware plugins (e.g., for VSCode) would improve usability and reduce friction.

\subsection{Future Work}
As PyGemini matures, several avenues for improvement and extension are anticipated. Some of these are proposals to meet current challenges, while others stem from topics that have not been given emphasis in this paper.

\subsubsection{Improving simulation capabilities}
The ability to perform synthetic reconstruction of both the environment and vessels---shown respectively in Figure \ref{fig:3dgs_from_rosbags} and Figure \ref{fig:3dgs_models}---is a key feature of PyGemini. Combined with ROS integration and sensor models with adjustable fidelity, this makes PyGemini particularly well-suited to support ongoing work in validating sensor models \cite{Vasstein:Hellinger}. This, would in turn, provide deeper insights into how simulation fidelity affects the runtime behaviour of autonomous systems, a crucial part in simulation-based assurance \cite{Glomsrud:Modular_assurance, Torben:contract_based_verification}.

However, as seen in Table \ref{tab:feature_comparison} there are several sensor and simulation aspects not yet supported by PyGemini. One of these are support for radar, which the previous Autoferry Gemini supported \cite{Vasstein_autoferry_gemini}. Since the existing lidar simulation is already based on radar equations (Appendix \ref{app:lidar_intensity}), this work is planned for future release.

Despite PyGemini's design allowing for faster communication and use of libraries that can optimize performance \cite{Ansel:Pytorch2, Lam:Numba}, this has yet to be shown. Hardware-accelerated modules, especially those we have seen for GPU-based sensor models from Stonefish \cite{grimaldi:stonefish} and Autoferry Gemini \cite{Vasstein_autoferry_gemini} could be introduced with minimal architectural disruption. This would optimize runtime for ray tracing tasks that currently run on CPU, such as PyGeminis lidar.

On the same topic, adding support for motion physics would be beneficial to make PyGemini suitable for control systems, and not just situational awareness. With existing motion physics libraries for Python that supports several marine vehicles both underwater and on the water surface \cite{Fossen:handbook}, integrating them with PyGemini should be feasible.

\subsubsection{Demonstrating Autonomy capabilities}
With reliable simulation validation systems in place in addition to an expanded simulation suite, a natural next step would be to incorporate closed-loop simulation with autonomy stacks for either HIL or SIL testing. This would enable verification of autonomy systems using simulated sensors that have been benchmarked against real-world operations, bridging the gap between physical testing and virtual validation. If PyGemini where to host the complete autonomy system as well, e.g. in a ROS node (Section \ref{sec:ROS_node}), it would give a demonstration of the ecosystems extensibility beyond its current priority of simulations. 

\subsubsection{Adding support for cybersecurity}
PyGemini has shown strong potential as a cybersecurity testbed, particularly in simulating adversarial physical environments. The ability to generate photorealistic simulations using generative models (as in Figure~\ref{fig:sd_pipeline}) already supports the creation of poisoned or resilient datasets. Building on this foundation, future work may include the simulation of perturbation-based attacks on sensor inputs, adversarial training tools, and configuration-based generation of cyber-physical threat models. These extensions would help meet growing research demands in maritime cybersecurity and adversarial AI \cite{Walter:AAI_maritime_testcases}, positioning PyGemini as a unique and versatile test arena in this domain.

\subsubsection{Improving development process}
Another promising direction is to further improve on PyGemini's CDD novelty. Currently, PyGemini’s test system can detect which components are affected by changes, but it offers little insight into the origin of bugs. By contextualizing component changes within the structure of a data pipeline---representing the flow of data through processors---developers could trace unexpected behavior back to specific processors. This would potentially improve debugging through data pipeline analysis. Since these pipelines often form tree-like hierarchies, they lend themselves well to automated analysis and visual inspection. Future improvements may include automated extraction of pipeline graphs from configurations (Figure \ref{fig:configuration_and_datapipeline}), their integration into test outputs, and mechanisms to compare pipelines across different versions to streamline validation and reduce unnecessary testing overhead.

Usability of configuration files also remains a critical area for improvement. As the system grows in scope, configuration files become more complex and prone to user error. Providing linting tools, along with editor support for features like autocompletion and inline documentation, would significantly reduce the learning curve and make the configuration-driven workflow more approachable, especially for users without deep software engineering experience.

\section{Conclusion}
In this paper, we introduced PyGemini, a modular, configuration-driven framework and library designed to unify simulation, autonomy, and visualization workflows for autonomous maritime systems. Through its entity-component-system (ECS) architecture and configuration-driven development (CDD) paradigm, PyGemini offers a flexible and extensible ecosystem for both researchers and practitioners working across diverse applications.

Our evaluation shows that PyGemini achieves meaningful progress in three key areas with use of CDD: integration, modularity, and maintainability. By decoupling core logic from deployment context, PyGemini supports seamless interoperability between ROS, Docker, and proprietary engines. This design enables developers to reuse the same codebase across different execution environments, reducing setup overhead and configuration drift. Alignment of embedded feature specifications directly in configuration files allows them to act as both deployment artifacts and test definitions, allowing PyGemini to serve as a shared language between domain experts and developers.

At the same time, this coupling of configurability and testability introduces new demands and possibilities for improving simulations and development---directions we highlight as important for future work. In addition, some integration gaps still remain, particularly where hardware constraints or vendor-specific systems resist full abstraction. GPU-dependent applications, for instance, may face compatibility issues that Docker alone cannot solve. Nonetheless, PyGemini’s data-oriented architecture mitigates some of these constraints by enabling selective execution: Lidar simulations can run independently of visualization pipelines, and sensor models can be swapped or disabled depending on hardware availability. Compared to existing solutions, this modularity allows PyGemini to scale more gracefully across use cases and hardware profiles.

Looking ahead, CDD should be general enough to be adapted for use beyond maritime domains. Its emphasis on modularity, data-centric design, and executable configurations makes it a promising candidate for cross-domain applications in aerospace, robotics, and smart infrastructure. By combining software engineering best practices with an open, extensible ecosystem, PyGemini lays a foundation for a more collaborative and transparent future in autonomous systems research.

As a final reminder. PyGemini is not meant to necessarily replace existing solutions, but to complement them. Its services and processors can be embedded within other ecosystems---whether game engines, simulators, control centers, or bespoke toolchains---providing a bridge between high-fidelity simulation and real-world autonomy workflows. By having PyGemini show what we could achieve if we where to combine these functionalities, we get a glimpse in how future systems could work as they would most certainly not be like the systems we rely on today in autonomous systems research.

\section*{Acknowledgment}
This work was supported in part by the Research Council of Norway through project 331921. The authors would also like to thank Tristan Elias Wolfram for his assistance in porting parts of the scenario generation from \cite{Fjellheim:Agents} to a more viable format for Python. 
\bibliographystyle{bibstyle.bst}
\bibliography{references}

\appendix
\section*{A: Simulated lidar intensity} \label{app:lidar_intensity}

\begin{figure}[ht]
    \centering
    \includegraphics[width=0.9\linewidth]{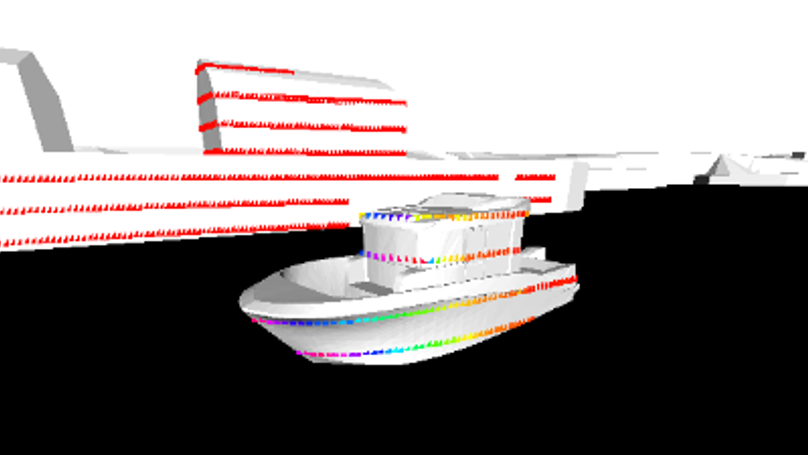}
    \caption{Lidar simulation with emitted intensity values and lidar textures for lambertian reflection. Intensity ranges from red to violet respectively illustrating high and low values. Textures ranges from black to white illustrating absorbent vs reflective material. Rays with intensity levels below an adjustable threshold value are dropped.}
    \label{fig:lidar_simulation}
\end{figure}

The received intensity \(I\) of each laser return is computed by combining geometrical optics, material reflectance models (BRDFs), and the radar equation. Below we summarize the main steps and formulae used.

\paragraph{Beamwidth and Laser Footprint}

The lidar lasers full beamwidth \(\theta_{\mathrm{bw}}\) can be approximated by the divergence of a Gaussian beam \cite[Eq.~5]{Alda:beam_divergence}:
\begin{equation}
  \theta_{\mathrm{bw}} \;=\; \frac{4\,\lambda}{\pi\,D_\mathrm{rx}},
\end{equation}
where \(\lambda\) is the laser wavelength and \(D_\mathrm{rx}\) is the receiver aperture diameter.

At a slant range \(R\), this divergence produces a circular laser footprint of area \cite[Eq.~17]{Wagner:static_radar_equation}:
\begin{equation}
  A_{\mathrm{fp}}(R) \;=\;
  \frac{\pi\,R^2\,\theta_{\mathrm{bw}}^2}{4}.
\end{equation}

\paragraph{Bidirectional Reflectance Distribution Function (BRDF)}

The interaction of the laser beam with surface materials is captured via a BRDF, \(f_r(\theta)\), which relates incident irradiance to reflected radiance at scattering angle \(\theta\). We implement two common models:

\paragraph{Lambertian BRDF}
For perfectly diffuse (Lambertian) surfaces \cite[Sec.~2]{OrenNayar:reflectance_models}:
\begin{equation}
  f_r^{\mathrm{Lambert}}(\rho) \;=\; \frac{\rho}{\pi},
\end{equation}
where \(\rho\in[0,1]\) is the surface reflectivity.

\paragraph{Oren–Nayar BRDF}
Rough surfaces are modeled using the Oren–Nayar generalisation of the Lambertian BRDF \cite[Eq.~30]{OrenNayar:reflectance_models} for lidars \cite[Eq.~7]{Carrea:oren_nayar_lidar_model}. Given reflectivity \(\rho\), RMS roughness \(\sigma\), and incident/scattering angle \(\theta\), we get:
\begin{align}
  A &= 1 \;-\; \frac{0.5\,\sigma^2}{\sigma^2 + 0.33}, \\
  B &= 0.45\,\frac{\sigma^2}{\sigma^2 + 0.09}, \\
  f_r^{\mathrm{O\-N}}(\rho,\sigma,\theta)
    &= f_r^{\mathrm{Lambert}}(\rho)\,\Bigl[A + B\,\sin\theta\,\tan\theta\Bigr].
\end{align}

\paragraph{Backscatter Cross Section}

The effective backscatter cross section \(\sigma_{\mathrm{bs}}\) (also called the radar cross section) relates the footprint area and BRDF to the fraction of power directed back toward the receiver \cite[Eq.~14]{Wagner:static_radar_equation}:
\begin{equation}
  \sigma_{\mathrm{bs}}(\theta, R)
  \;=\;
  4\pi\,\cos\theta \;A_{\mathrm{fp}}(R)\;f_r(\theta)\,.
\end{equation}

\paragraph{Static Radar Equation}

Finally, the received signal intensity \(I\) at range \(R\) follows the static radar (lidar) equation \cite[Eq.~24]{Wagner:static_radar_equation}:

\begin{equation}
    I(R)=\frac{P_{t} D_\mathrm{rx}^{2}}{4 \pi R^{4} \theta_{\mathrm{bw}}^2} \eta_{\mathrm{opt}} \eta_{\mathrm{atm}} \sigma_{\mathrm{bs}}
\end{equation}

where
\begin{itemize}
  \item \(P_t\) is the transmitted power of the lidar,
  \item \(\eta_{\mathrm{opt}}\) is the optical efficiency of the system,
  \item \(\eta_{\mathrm{atm}}\) is the efficiency of two-way atmospheric attenuation
  \item all other symbols are as defined above.
\end{itemize}

Combining these yields per-ray intensity returns that account for divergence, range attenuation, atmospheric losses, and material reflectance. The latter (\(\rho\) and \(\sigma\)) are expressed as texture components attached to scene entities as shown in Figure \ref{fig:lidar_simulation}.

\end{document}